\documentclass[default,iicol]{sn-jnl}

\usepackage{graphicx}%
\usepackage{multirow}%
\usepackage{amsmath,amssymb,amsfonts}%
\usepackage{amsthm}%
\usepackage{mathrsfs}%
\usepackage[title]{appendix}%
\usepackage{xcolor}%
\usepackage{textcomp}%
\usepackage{manyfoot}%
\usepackage{booktabs}%
\usepackage{algorithm}%
\usepackage{algorithmicx}%
\usepackage{algpseudocode}%
\usepackage{listings}%

\usepackage{subcaption}
\usepackage{array}
\usepackage{multirow}
\usepackage{hhline}
\usepackage{amssymb}

\theoremstyle{thmstyleone}%
\theoremstyle{thmstyletwo}%

\theoremstyle{thmstylethree}%

\begin{document}

\title[Image Generation and Training Strategy for Deep Document Forgery Detection]{Image Generation and Learning Strategy \linebreak for Deep Document Forgery Detection}


\author*[1,2]{\fnm{Yamato} \sur{Okamoto}}\email{okamoto.yamato.w15@kyoto-u.jp}
\author[3]{\fnm{Osada} \sur{Genki}}
\author[4]{\linebreak \fnm{Iu} \sur{Yahiro}}
\author[5]{\fnm{Rintaro} \sur{Hasegawa}}
\author[3]{\fnm{Peifei} \sur{Zhu}}
\author[3]{\fnm{Hirokatsu} \sur{Kataoka}}

\affil[1]{\orgname{NAVER Cloud Corp}}
\affil[2]{\orgname{WORKS MOBILE JAPAN Corp}}
\affil[3]{\orgname{LY Corporation}}
\affil[4]{\orgname{The University of Tokyo}}
\affil[5]{\orgname{Keio University}}


\abstract{
In recent years, document processing has flourished and brought numerous benefits. However, there has been a significant rise in reported cases of forged document images. Specifically, recent advancements in deep neural network (DNN) methods for generative tasks may amplify the threat of document forgery.
Traditional approaches for forged document images created by prevalent copy--move methods are unsuitable against those created by DNN--based methods, as we have verified.
To address this issue, we construct a training dataset of document forgery images, named FD-VIED, by emulating possible attacks, such as text addition, removal, and replacement with recent DNN--methods. Additionally, we introduce an effective pre-training approach through self-supervised learning with both natural images and document images.
In our experiments, we demonstrate that our approach enhances detection performance.}

\keywords{Document forgery detection, Document synthesis, Document forensics.}



\maketitle

\section{Introduction}\label{sec1}

\begin{figure}[htb]
\begin{center}
    \begin{minipage}[b]{3.5cm}
        \centering
        \includegraphics[width=3.5cm]{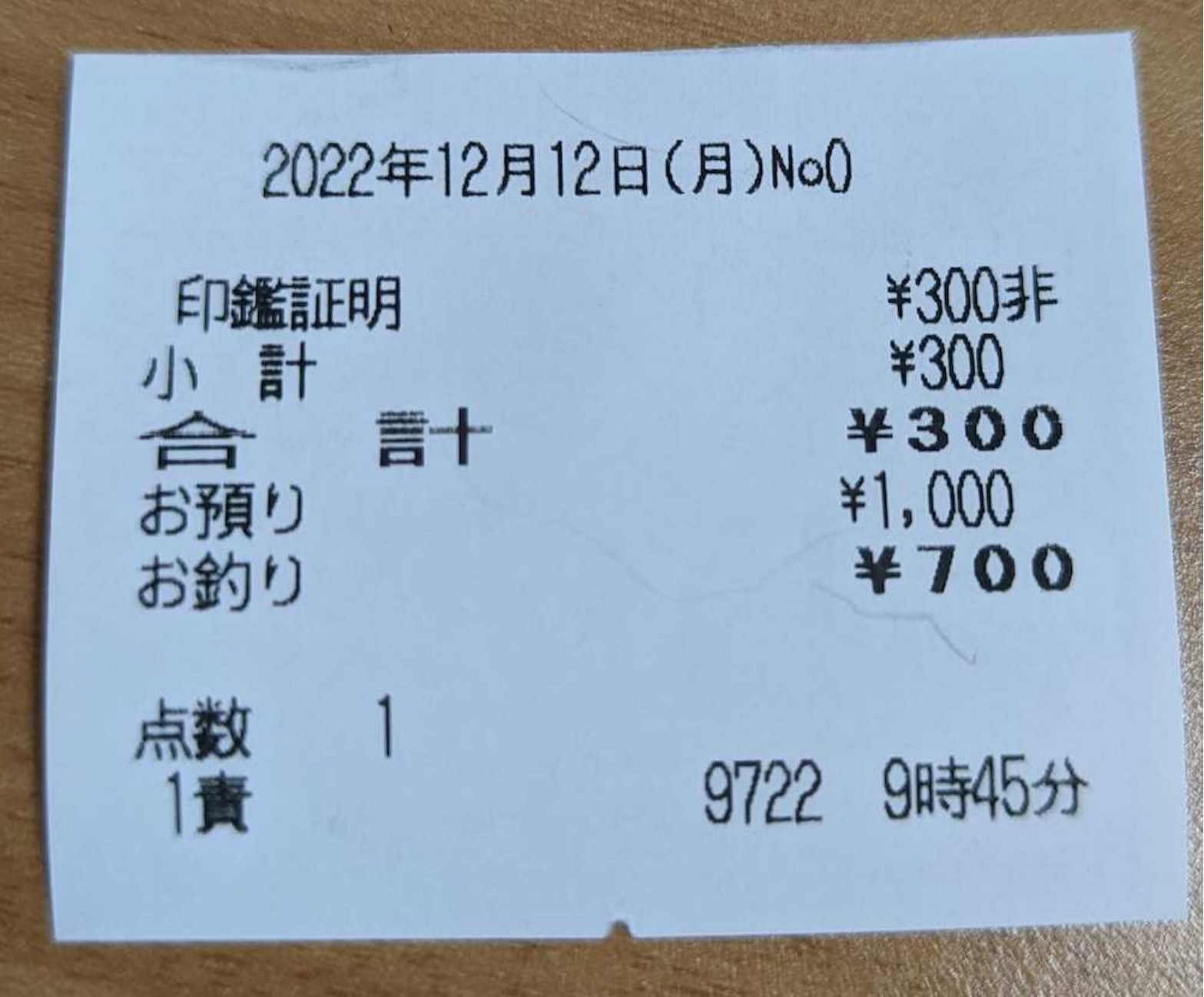}
        \subcaption{Authentic image}
    \end{minipage}
    \begin{minipage}[b]{3.5cm}
        \centering
        \includegraphics[width=3.5cm]{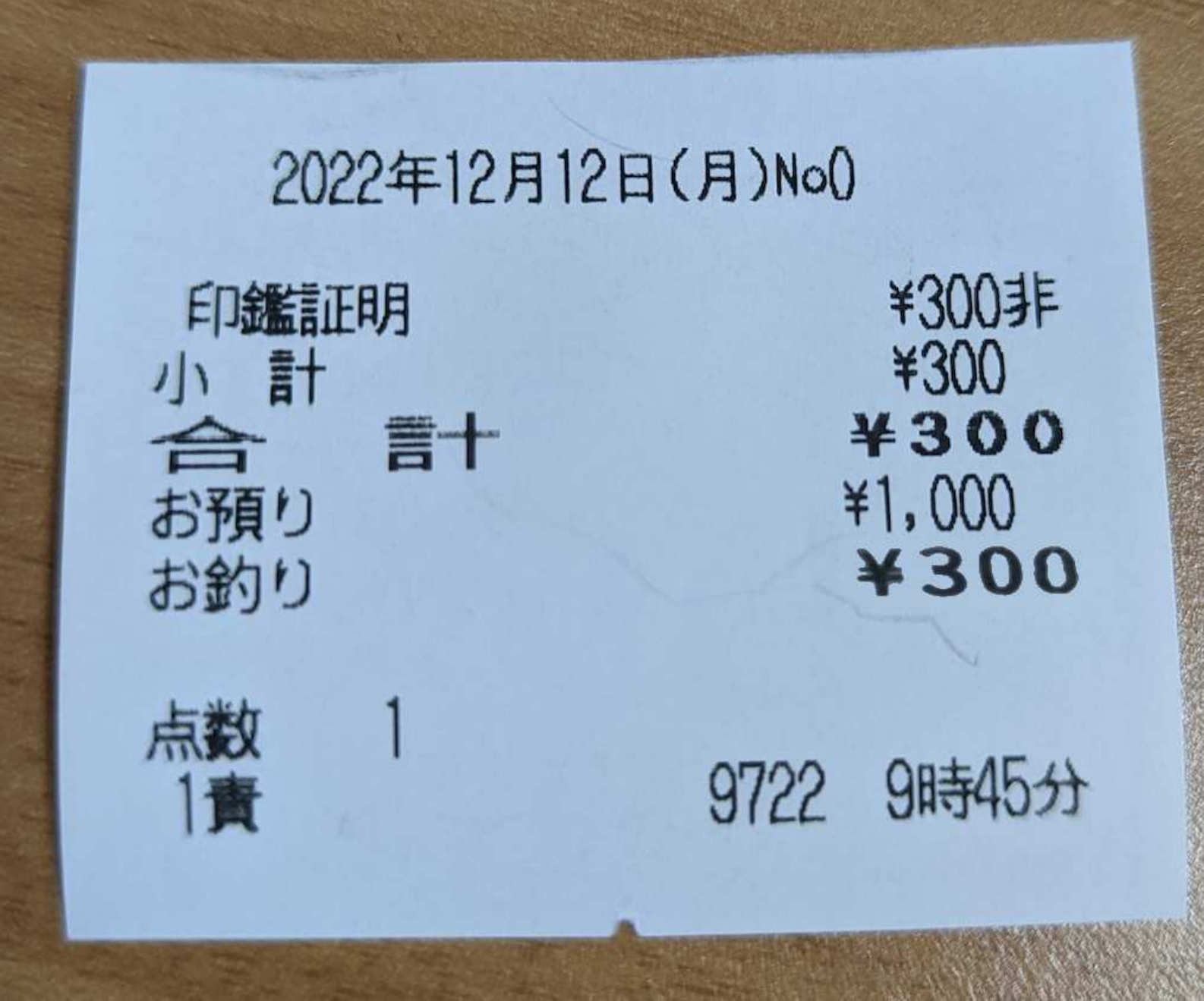}
        \subcaption{Forged image}
    \end{minipage}
    \begin{minipage}[b]{3.5cm}
        \centering
        \includegraphics[width=3.5cm]{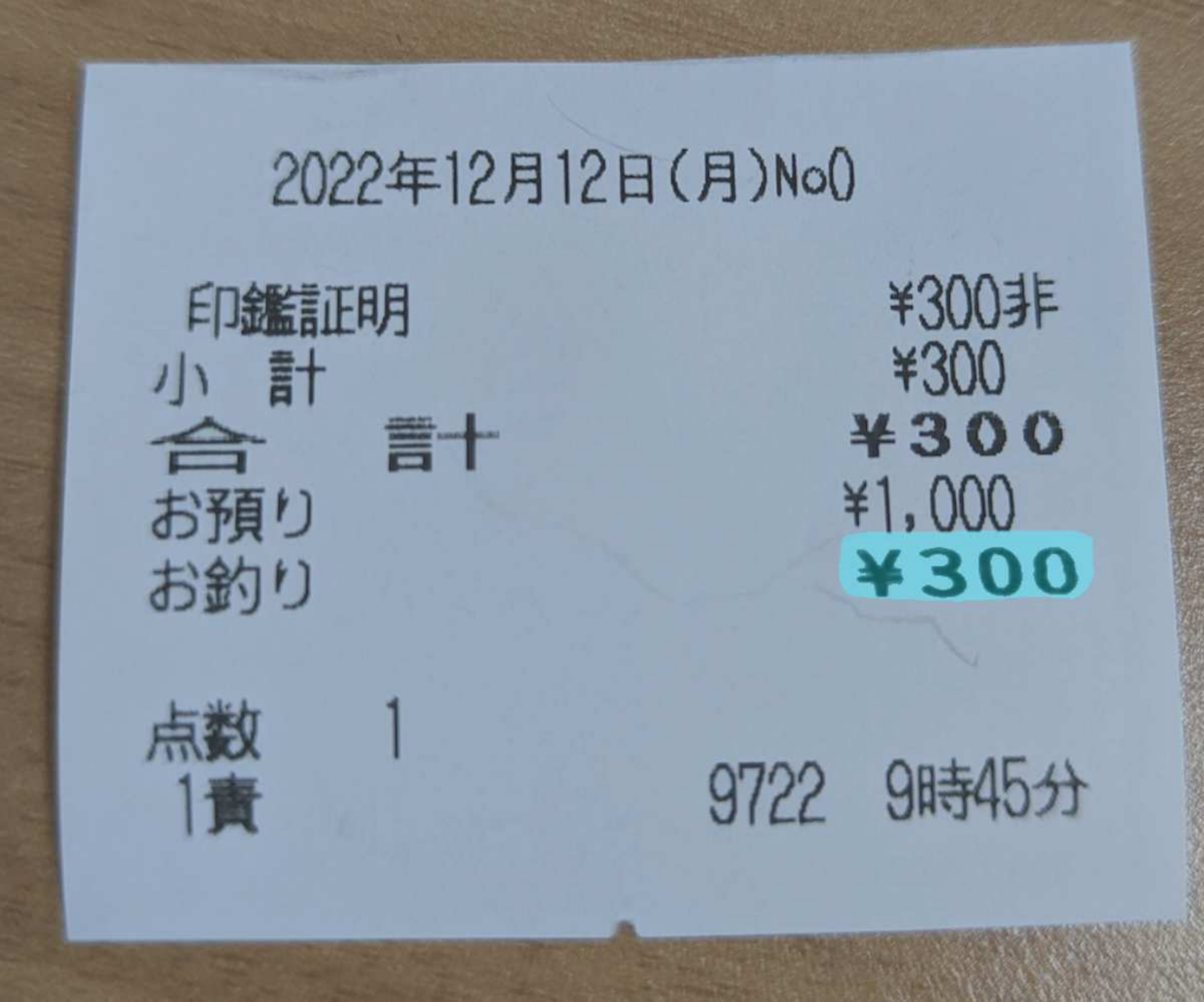}
        \subcaption{Detection result}
    \end{minipage}
\caption{Distinguishing forged images created by DNN--based methods from authentic images is challenging even for humans. Our proposed detection method effectively identifies the forged areas (highlighted in blue).}
\label{fig:sample_result}
\end{center}
\end{figure}

Document processing that converts analog documents into a digital format is now flourishing. Document handling tasks such as invoice processing, loan credit checks, and insurance applications are becoming digitized, thereby improving operational efficiency. Moreover, the utilization of digitized data will enhance competitiveness and improve the quality of services. Thus, the introduction of document processing is now being accelerated in a wide range of industries.

However, despite the positive aspects of document processing, abuses in document processing are often reported. As an example, the forging of text in the document image, which is the input for document processing, can disguise an identity or have other fraudulent incentives. Moreover, such forgery can be achieved with ease by using readily available image-editing software, such as smartphone photo-editing apps, requiring only minimal effort.

This problem is due to the vulnerability of the optical character reader (OCR), which is involved in the first step of document processing. OCR is a technology that recognizes text on imaged documents. Its main focus is reading text faithfully, not robustness against forgery. Therefore, OCR inadvertently reads the text altered by forgery and fails to read the text that has been removed by forgery.

To make matters worse, when used improperly, the recent advancement in deep neural network (DNN) for generative tasks may amplify the threat of document forgery.
Unlike traditional forgery (referred to as the copy--move method in this paper), modern DNN--based editing technologies introduces sophisticated forgeries that are challenging to detect.
Inpainting, a technique that complements any part of an image without discomfort, has become increasingly sophisticated, and the technique of editing text in images called scene text editing (STE) is no longer distinguishable by the human eye.
Even systems that rely on human inspection, as commonly employed in financial institutions, may face challenges in identifying such advanced methods of forgery. The reality and danger of this scenario have been demonstrated by the problems resulting from deepfakes based on a similar generative DNN model.

In this work, we address the detection of such document forgeries. In our experiments, we observed that the models, designed to detect document forgeries created by prevalent copy--move methods, perform poorly for those created by modern DNN--based methods (refer to Section \ref{section:finetune}). 
To overcome this, we focus on the following aspects.
\begin{itemize}
    \item {
    \emph{Building a dataset with various forgeries.}   
    We develop a DNN--based document image editing method that allows more elaborate forgery. Specifically, we use the inpainting method to remove text and scene text editing to add and rewrite the text. Expanding a training dataset by such forgery images, the detection model becomes adept at capturing the inconsistencies present within them.
    }
    \item {
    \emph{Exploring suitable pre-training strategies.}
    The efficacy of pre-training has been well-established across diverse machine learning domains. In this work, we demonstrate the importance of pre-training in the context of document forgery detection. 
    Notably, while our target involves document images, we found that pre-training with not only document images but also with ImageNet, a dataset consisting of natural images, has a huge effect.
    }
\end{itemize}
With the above efforts, we achieve a method of defending against the threat of forged documents created by modern DNN--based editing technology (An example is shown in Fig.\ \ref{fig:sample_result}). It would be a baseline for future research in this field and our discoveries are expected to be useful for the future development of this field.

\section{Related Work}

\subsection{Forgery Detection}
Research on forgery detection can be categorized based on the types of images it targets.

\subsubsection{Document forgery detection}
This is the research area that our study focuses on. Francisco \cite{Categorization} defines forgery patterns in document images, including the copy--move method, and compiles visual features and clues for detection. To detect such forgeries, several studies have been conducted. For instance, deep neural networks have been employed to determine the presence of forgery and to localize it \cite{XuWenbo}. Moreover, DNN--based approaches have explored network architectures leveraging attention mechanisms \cite{e24010118, 10063499} and graph neural networks \cite{10.1007/978-3-031-09037-0_22}. 

Such studies use private datasets or publicly available datasets \cite{hal-02316349, 8090394, tianchi}. It is important to highlight that in both cases forgeries are predominantly created through copy--move or splicing techniques, not DNN--based techniques. This has led to insufficiency in validating forgeries created using recent generative DNN--based techniques. In our paper, we redefine forgery patterns by integrating the perspective of forgery operations using DNN--based methods, and construct a comprehensive dataset for further verification.

\subsubsection{Forgery detection for natural images}
Forgery detection in the domain of natural images has become an active area of research.
Segmentation-based methods have typically been used in recent times; e.g., NEDB-Net \cite{nedb-net}, MVSS-Net \cite{Chen_2021_ICCV}, and RRU-Net \cite{bi2019rru}.

Forgery detection for scene text images \cite{10.1007/978-3-031-19815-1_13} can be considered a related yet distinct research field. Its objective is to detect forged text within natural images, such as signage. While our study shares similarities in terms of text manipulation, the difference lies in the target being document images. 
When the focus is on documents, it is expected that the clues for detecting forgery will differ, such as the availability of color features, for example.

\subsubsection{Facial forgery detection}
Facial forgery detection is extensively researched \cite{zhou2017two, yang2019exposing, li2020face} because there have been various works on facial forgery generation; e.g., \cite{karras2019style, gonzalez2018facial, choi2018stargan, perov2020deepfacelab, liu2019stgan} and these facial forgery has already become a social problem.
Facial forgery detection task is primarily designed as a binary classification of forgery or authenticity, rather than localization. From this perspective, we believe that document forgery detection is different from facial forgery detection.

\subsection{Image Editing}
Image editing techniques have primarily advanced in the realm of natural images rather than in the context of document images. Therefore, in this study, we incorporate image editing techniques for the domain of natural images to construct a dataset of forged document images. In the following, we focus on and list related DNN--based techniques.

Inpainting is a DNN--based technique used to generate completions for arbitrary regions in an image.
We apply this technique to document images to remove text and replace it with a forged background.
Several inpainting methods, such as DeepFillv1 \cite{yu2018generative}, Global\&Local \cite{iizuka2017globally}, PConv \cite{liu2018image}, and AOT-GAN \cite{zeng2022aggregated}, have been proposed, and among them, we select DeepFillv2 \cite{yu2019free} because of its stable performance for document images.

Scene text editing (STE) is a method that converts the text within natural images into other text while maintaining the original font style, and we apply this to generate forged text on document images.
STE was introduced by SRNet \cite{wu2019editing}, and since then, many methods, such as  RewriteNet \cite{lee2022rewrite}, MOSTEL \cite{qu2022exploring}, STEFANN \cite{roy2020stefann}, and De-rendering \cite{shimoda2021rendering}, have been proposed.
In this work, we choose SRNet because it enables both model learning and inference (i.e., image generation) in an end-to-end manner.

It should be noted that Vision-Text-Layout Transformer, a model that handles text, image, and layout modalities, achieving document generation and document editing has been proposed \cite{Tang_2023_CVPR}. However, as the trained model is not currently available, we did not validate this in our work.

\newcolumntype{K}{>{\centering\arraybackslash}m{8.6em}}
\begin{figure*}[tbp]
    \centering    
    \begin{tabular}{c*3{K}}
        \toprule
         & Input image & \multicolumn{2}{c}{OCR output}  \\ 
        \midrule
        \begin{tabular}{c} Original\\ image 
        \end{tabular}
        & \includegraphics[width=3.2cm]{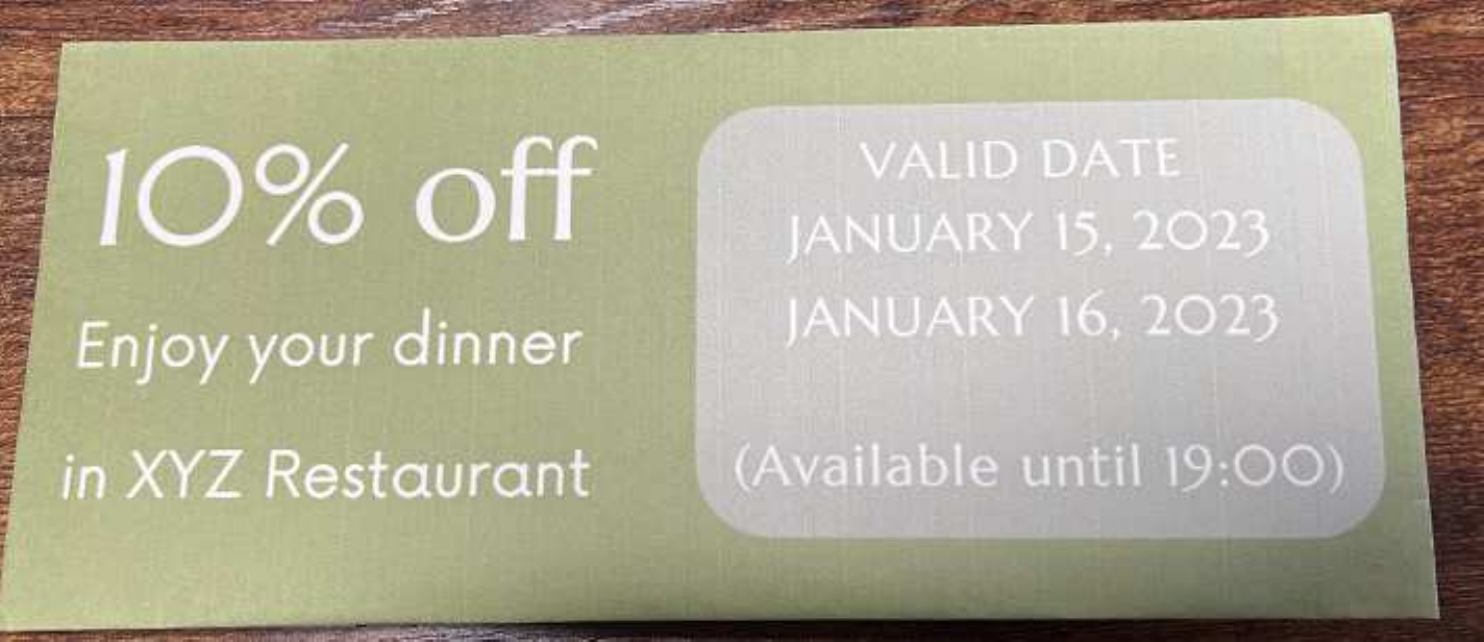} &
        \includegraphics[width=3.2cm]{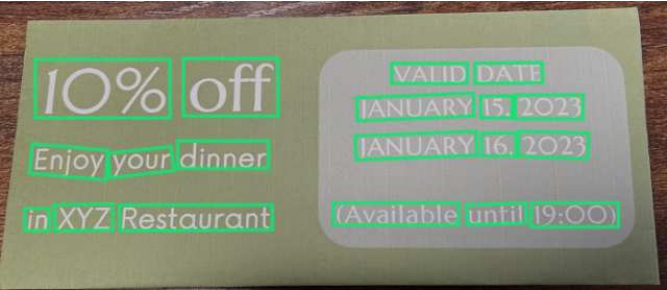} 
        &
        \includegraphics[width=3.2cm]{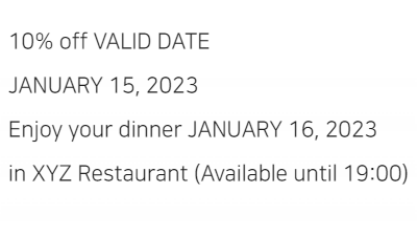}   
        \\ 
        \begin{tabular}{c} Text\\ removal 
        \end{tabular} & \includegraphics[width=3.2cm]{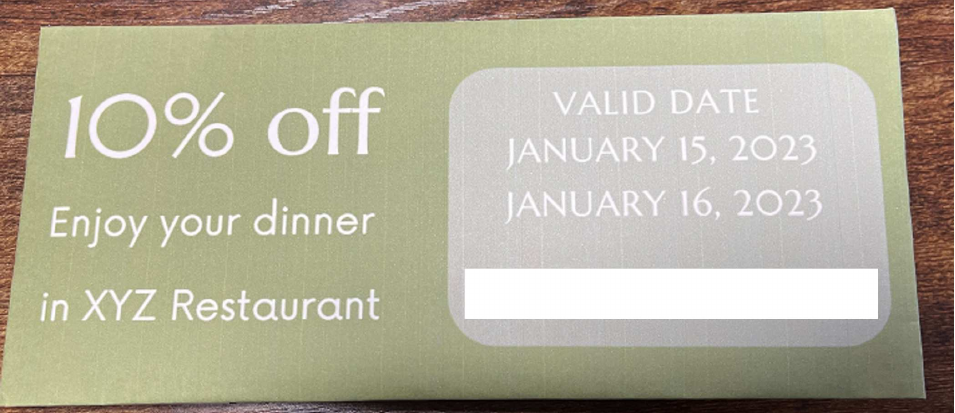} &
        \includegraphics[width=3.2cm]{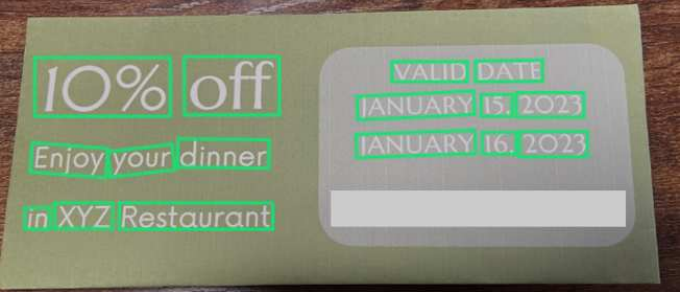} 
        &
        \includegraphics[width=3.2cm]{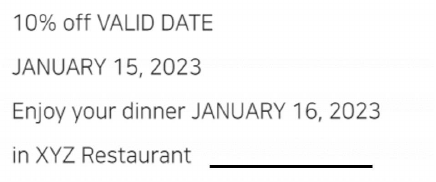} 
        \\
        \begin{tabular}{c} Text\\ addition 
        \end{tabular} & \includegraphics[width=3.2cm]{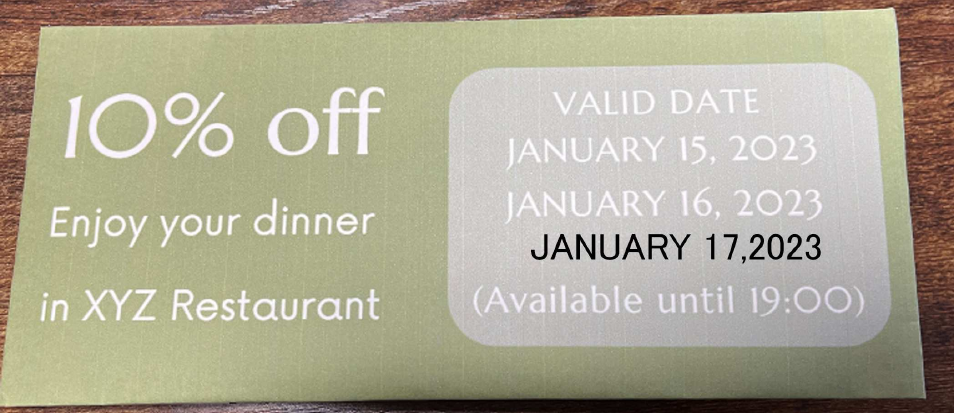} &
        \includegraphics[width=3.2cm]{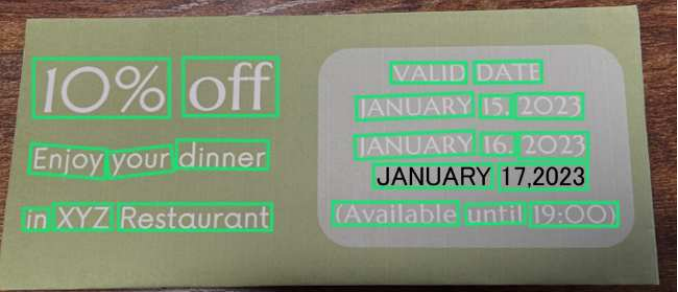} 
        &
        \includegraphics[width=3.2cm]{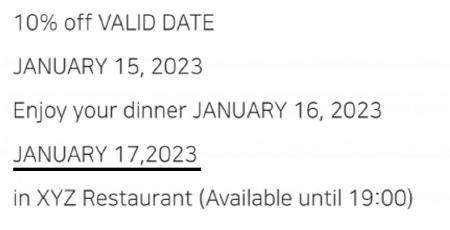}   
        \\
        \begin{tabular}{c} Text\\ replacement 
        \end{tabular} 
        & \includegraphics[width=3.2cm]{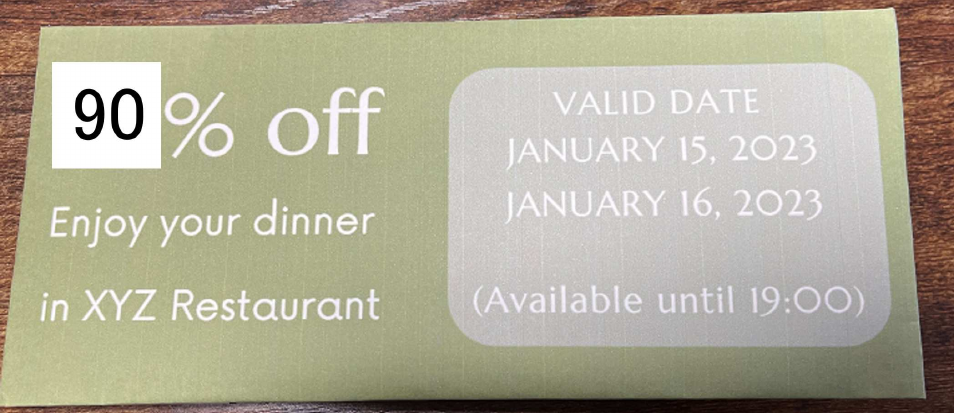} &
        \includegraphics[width=3.2cm]{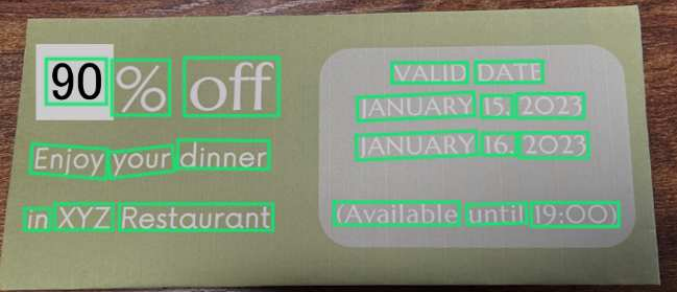} 
        &
        \includegraphics[width=3.2cm]{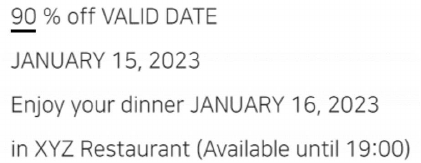} 
        \\
        \bottomrule 
    \end{tabular}
    \caption{Examples of controlling the OCR output arbitrarily through the forging of document images. 
    Even image editing that is easy for the human eye to detect can change the OCR output. 
    This poses a vulnerability to a document processing system.}
    \label{fig:preliminary}
\end{figure*}

\subsection{Self-supervised Pre-training}
\label{sec:document_understanding}
Pre-training is effective in improving the performance of a machine learning model.
In particular, self-supervised-learning pre-training methods such as SimCLR \cite{pmlr-v119-chen20j} and MoCo \cite{He_2020_CVPR} have achieved excellent performance.
More recently, as pre-training methods specialized for document understanding tasks, TrOCR \cite{li2023trocr} and DocFormer \cite{Appalaraju_2021_ICCV} have been proposed. Despite the limitation of requiring both document images and text annotations, the effectiveness of these methods has been confirmed for the OCR, layout recognition, and document classification tasks.
Additionally, Afzal et al. \cite{8270080} reported that pre-training with document images is more effective in document understanding tasks than pre-training with natural images such as those in ImageNet \cite{ImageNet}, and some document image datasets have been proposed \cite{li2020docbank}.

Although various pre-training methods and datasets are available, an optimal pre-training method for document forgery detection has not yet been studied. In this work, we explore effective pre-training methods step by step by conducting carefully designed experiments.

\section{Preliminary Study}
\label{Preliminary_Study}
In this section, we explain the vulnerability of the OCR for forged document images. The output of the OCR is easily controlled by editing input document images, even if the editing is easily detected by the human eye owing to inconsistent background colors and mismatched fonts.
Figure\ \ref{fig:preliminary} shows examples. 
In the first example, blanking out some text from the original document image can result in the removal of the corresponding text from the OCR output. 
In the second example, adding text on the image can insert additional text in OCR output.
In the third example, adding arbitrary text after blanking out some text areas makes OCR output different text from the original text.
In the context of the OCR's purpose to convert text contained in an image into digital symbols, these OCR behavior described above cannot be said to be wrong. Addressing this vulnerability requires not an improvement in the accuracy of OCR but a totally new function that detects forged document images.

\begin{figure*}[htbp]
\begin{center}
\includegraphics[width=12cm]{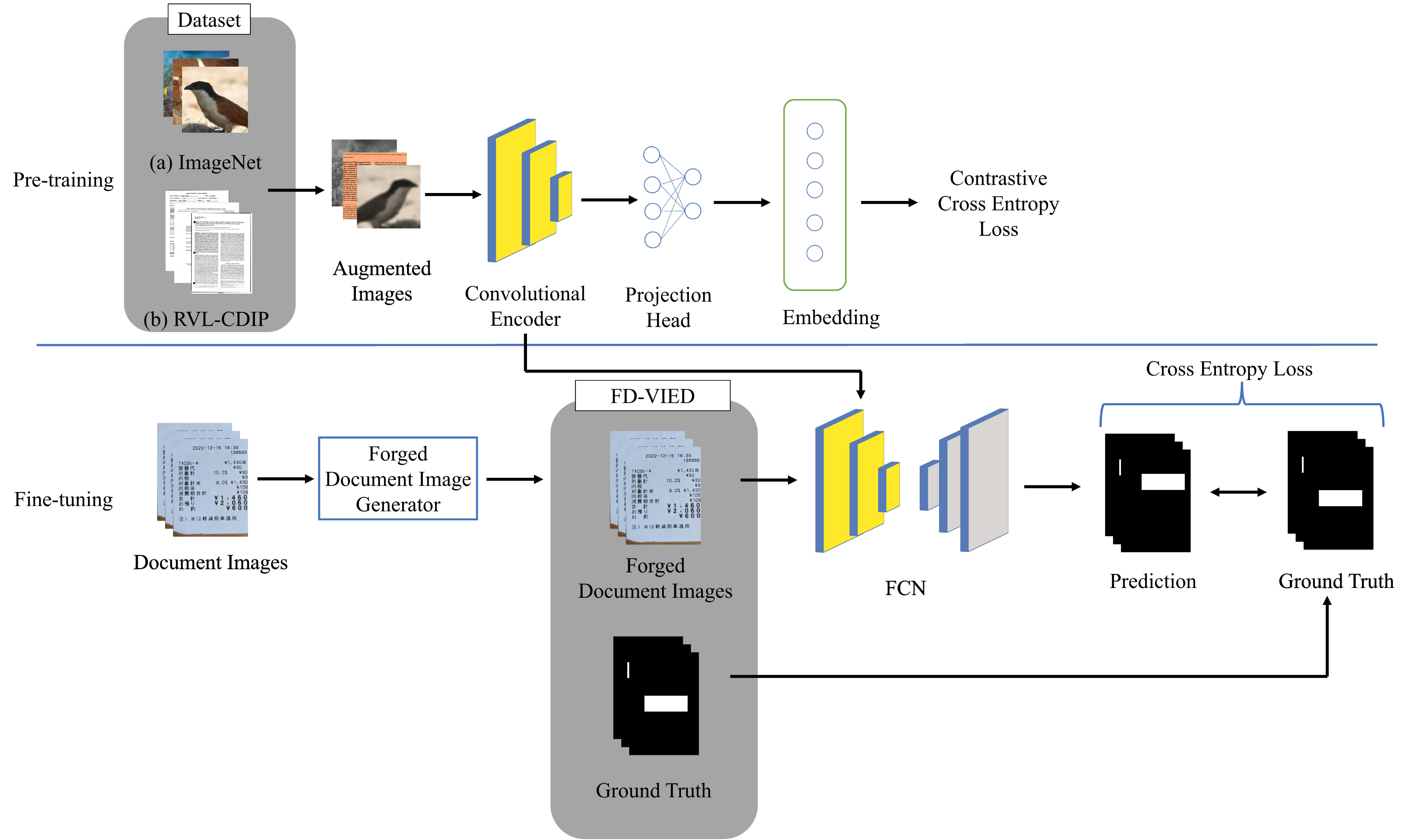}
\caption{Overview of building a document forgery detection model. In the pre-training stage, we employ self-supervised learning (SimCLR) with available public datasets. In the fine-tuning stage, we train a semantic segmentation model (FCN) through supervised learning on our forged document image dataset (FD-VIED).}
\label{fig:overview}
\end{center}
\end{figure*}

\section{Document Forgery Detection Model}
\label{sec:model_training}
Figure\ \ref{fig:overview} is an overview of the building of the document forgery detection model. 
For practical use, rather than making a binary judgment of whether a document is authentic or forged, we aim to localize the forged areas in images. Therefore, we use a semantic segmentation model of two-class (authentic/forged) pixel-wise classification, which is mainstream in traditional forgery detection for document images and natural images.

In the pre-training stage, we train the model through self-supervised learning on existing available public datasets to learn useful features for the downstream task. In the fine-tuning stage, we use the pre-trained model as the backbone and train document forgery detection models through supervised learning with forged document images. 

As described in Section~\ref{sec:document_understanding}, pre-training is expected to be effective for document forgery detection as well. However, in contrast with the previous tasks, there has still been no study showing the optimal pre-training method for document forgery detection. 
To explore the optimal pre-training approach, we consider that the optimal dataset for pre-training is different depending on the downstream task. The input images for document forgery detection are document images, and pre-training with document images might therefore be appropriate. Additionally, because artifacts and color differences are expected to be useful for document forgery detection, pre-training with natural images that include rich colors and textures may also be appropriate. 
Therefore, we use SimCLR, which has no limitations regarding input images, and evaluate models pre-trained on a variety of datasets. Additionally, as studied on SimCLR, the effectiveness largely changes with hyperparameter tuning and data augmentation, and we explore better configurations.

\newcommand{\addpic}{\includegraphics[width=6em]{example-image}}
\newcolumntype{C}{>{\centering\arraybackslash}m{5.5em}}
\begin{table*}[htbp]
    \captionsetup{singlelinecheck=off, justification=raggedright}
    \caption{Sample images of FD-VIED.}
    \label{tbl:samples}
    \begin{tabular}{lCCCC|C}    
        \toprule
        Editing method & Text removal & Text replacement & Text addition & Background addition & Original image \\
        \midrule
        Copy--move   &  \includegraphics[width=5.5em, trim={0 0 0 2em},clip]{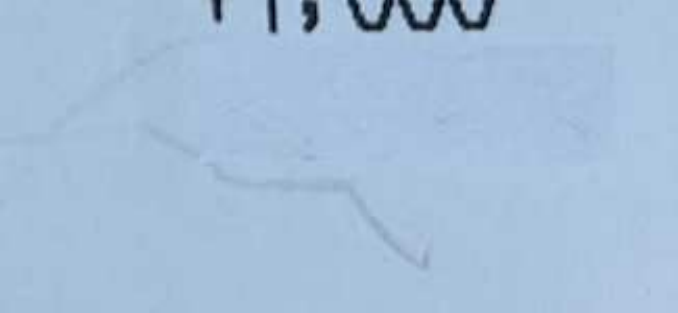} & \includegraphics[width=5.5em, trim={0 0 0 2em},clip]{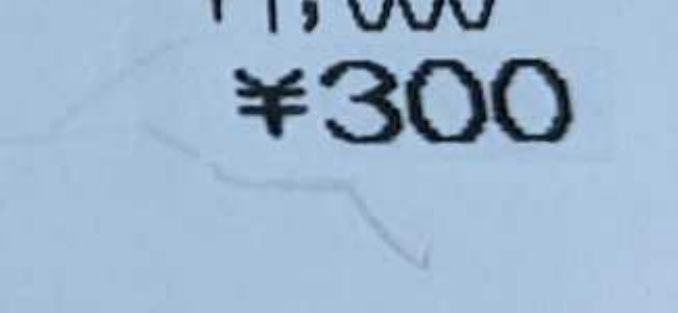} & \includegraphics[width=5.5em, trim={0 0 0 2em},clip]{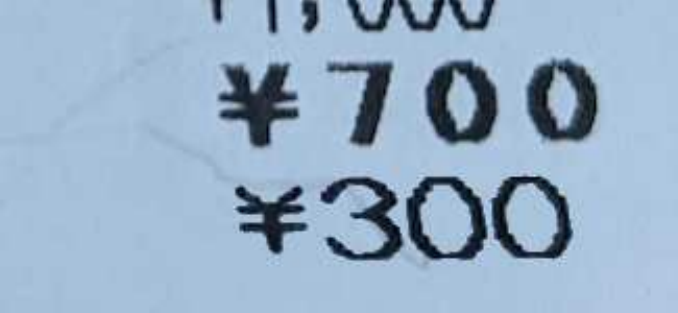} & \includegraphics[width=5.5em, trim={0 0 0 2em},clip]{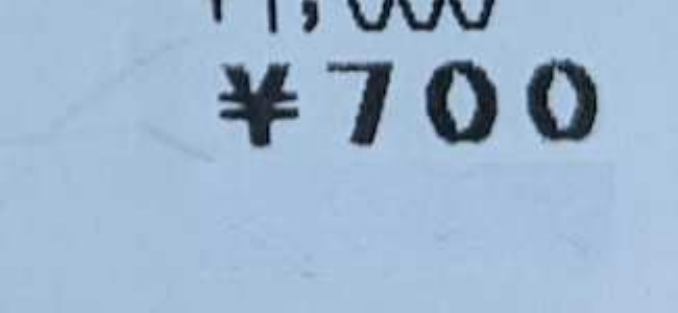} & 
        \includegraphics[width=5.5em, trim={0 0 0 2em},clip]{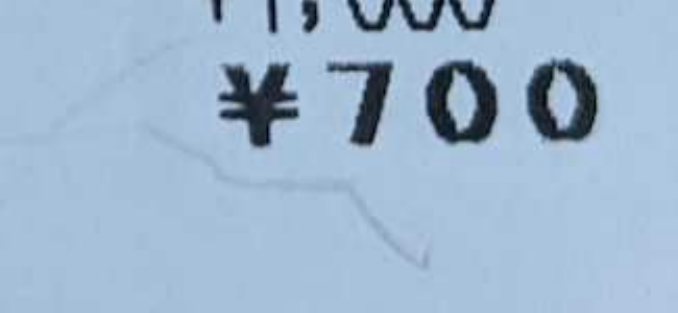} \\
        DNN--based & \includegraphics[width=5.5em, trim={0 0 0 2em},clip]{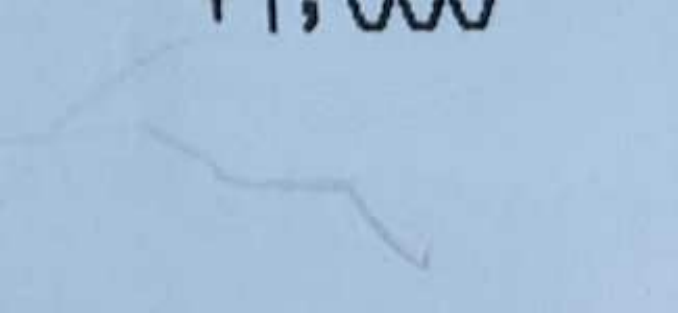}  & \includegraphics[width=5.5em, trim={0 0 0 2em},clip]{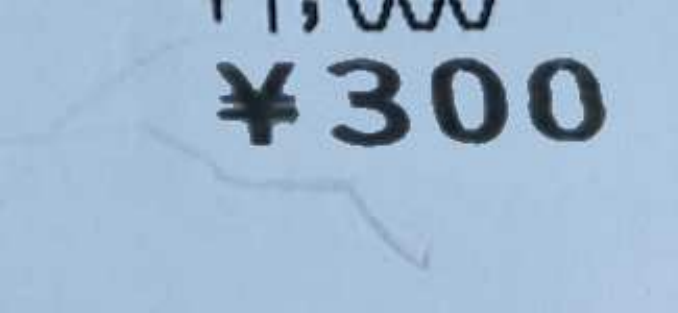}  & \includegraphics[width=5.5em, trim={0 0 0 2em},clip]{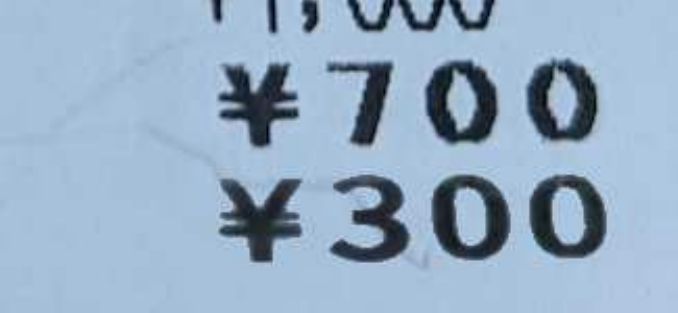} & \includegraphics[width=5.5em, trim={0 0 0 2em},clip]{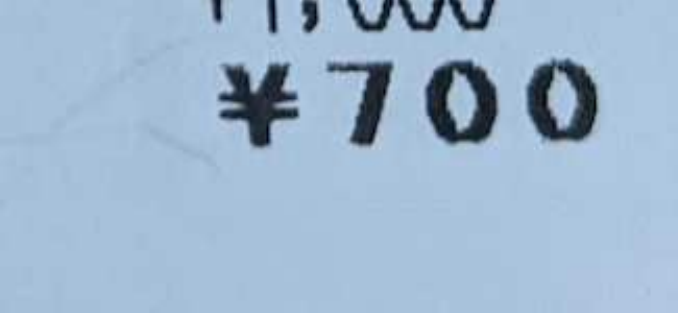} & 
        \includegraphics[width=5.5em, trim={0 0 0 2em},clip]{figs/receipt_imgs_0.pdf} \\
    \bottomrule 
    \end{tabular}
    \captionsetup{justification=centering}
\end{table*} 
\begin{figure*}[htbp]
    \begin{minipage}[b]{0.5\linewidth}
        \centering
        \includegraphics[width=1.0\linewidth]{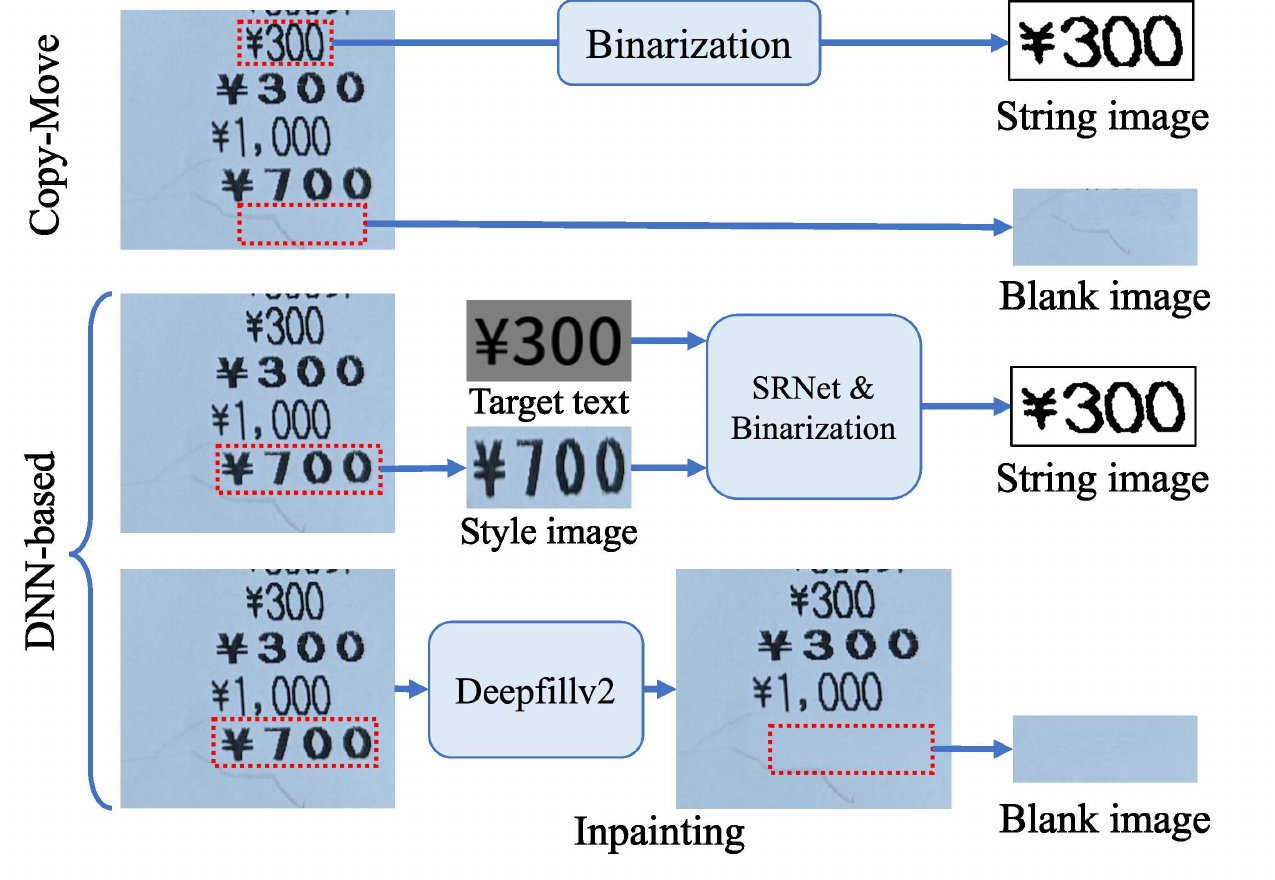}
        \subcaption{Editing methods.}
        \label{fig:flow_a}
    \end{minipage}%
    \begin{minipage}[b]{0.5\linewidth}
        \centering
        \includegraphics[width=1.0\linewidth]{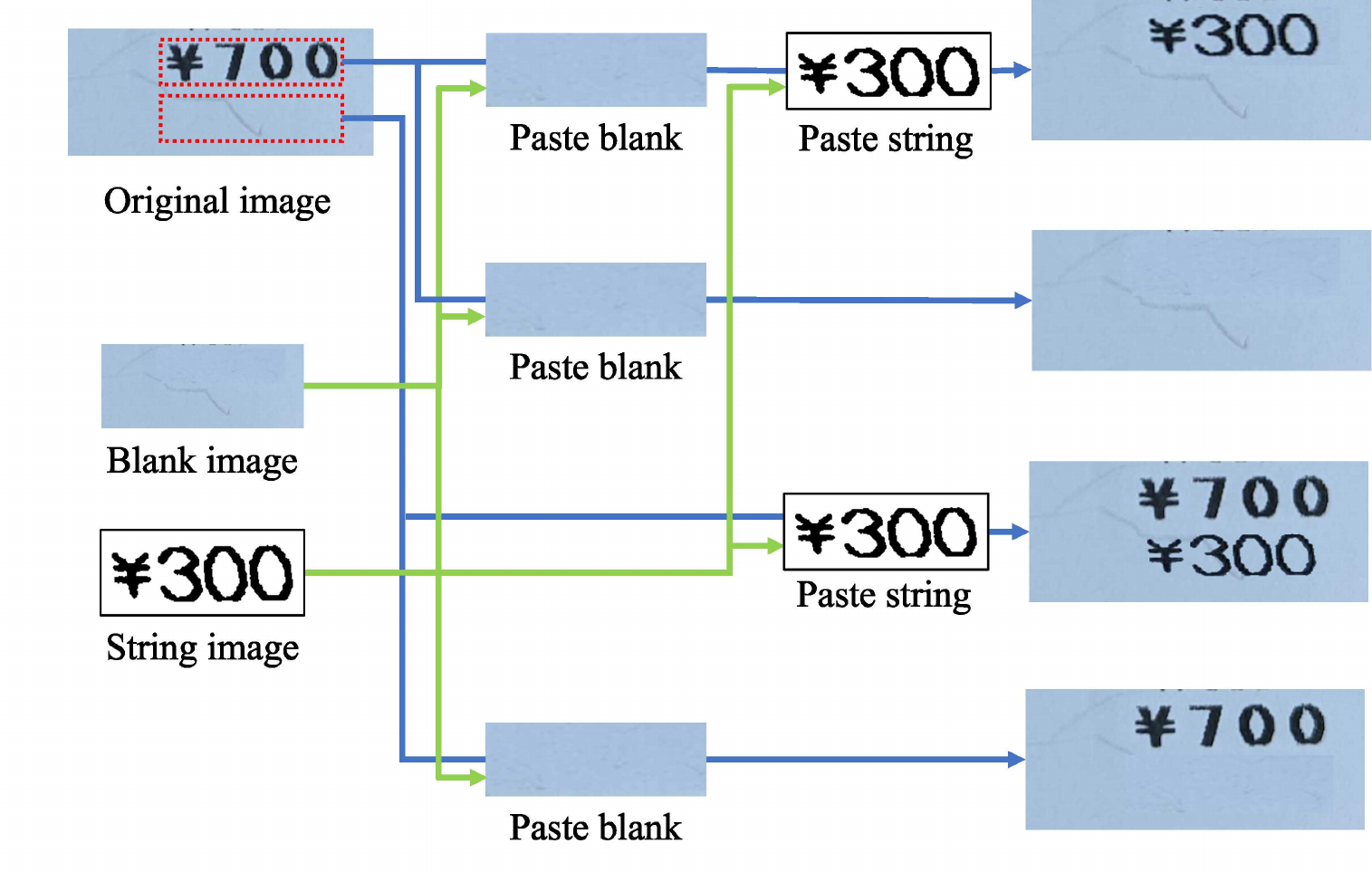}
        \subcaption{Editing patterns.}
        \label{fig:flow_b}
    \end{minipage}
    \captionsetup{singlelinecheck=off, justification=raggedright}
    \caption{
    Image generation overview: Editing the price text area or adjacent margin area based on the OCR result.
    }
    \captionsetup{justification=centering}
    \label{}
\end{figure*}

\section{Dataset Construction: FD-VIED}
\label{sec:dataset}
Creating a dataset for forgery document detection is necessary because real documents submitted by users often have restrictions on their secondary use due to privacy. 
Additionally, the number of forged document images that can be collected is not large enough to train a specific model.
Therefore, we have developed our dataset, which we call the Forged Document Images with Various Image Editing Dataset (FD-VIED).

We define four patterns of document tampering: text replacement, text removal, text addition, and background addition. Please note that background addition, while unlikely to occur in practical scenarios, is included to enhance comprehensiveness, as it is theoretically possible.
Each pattern can be realized by text pasting operation, background pasting operation, and a combination of both. On this basis, we implement each operation using the \emph{copy--move} and \emph{DNN--based} methods, respectively.
The sample images that we made are shown in Table \ref{tbl:samples}.

\subsection{Editing Methods}
We describe two editing methods. The process is depicted in Fig. \ref{fig:flow_a}.
\paragraph{Copy--move method.}
In the research field of forgery detection, copying a part of an image and pasting it into the same image is called the copy--move method, whereas pasting a part of an image into a different image is called splicing. 
Copy--move editing is easier for adversaries to perform, and the images forged using this method are harder to detect. We therefore adopt copy--move editing.
\paragraph{DNN--based methods.}
As for DNN--based text generation, we use STE, specifically SRNet \cite{wu2019editing} trained on images generated using SynthText \cite{Gupta16}. STE can generate realistic text images with fonts that match the fonts of surrounding text without the need for manual adjustment.
To generate background images, we use inpainting. Specifically, we use DeepFillv2 \cite{yu2019free} provided in \cite{mmediting2022}. In contrast with the copy--move method which often generates artifacts, inpainting generates more natural images.

\subsection{Editing Patterns}
We describe the editing patterns. An overview of the generation is shown in Fig.\ \ref{fig:flow_b}.
\paragraph{Text removal.}
In the copy--move method, we arbitrarily select a specific blank area (i.e., an area with no text) in the document image and copy it, and then overlay the copied blank image onto the text area we want to remove.
In the DNN--based methods, we mask the target text area and generate a blank image on the masked area from scratch through inpainting.
\paragraph{Text replacement.}
In the copy--move method, we copy a specific area with text and paste it on the target text area.
In the DNN--based methods, we remove the text from the target area in the same manner as the text removal, and then paste a \emph{string image} on the area. To clarify, a \emph{string image} means a text image with a transparent background, obtained by binarizing the text image generated from scratch with SRNet. SRNet can generate sophisticated text images, but when it comes to the background area, we identified that there are sometimes artifacts and unnatural to the human eye. Our method with binarization allows us to discard such unfavorable backgrounds and generate intricately forged images.
\paragraph{Text addition.}
In both the copy--move and DNN--based methods, we paste a \emph{string image} on the target area where any text does not exist. In copy--move method, string images are obtained by binarizing text images copied from the same document image.
\paragraph{Background addition.}
In both the copy--move and DNN--based methods, background addition is almost the same as text removal and differs from it merely in that text does not exist in the target area.

\subsection{Statistics of the FD-VIED}
Table~\ref{tabale:statistics} gives the statistics.
FD-VIED consists of 46,874 training images and 5,973 testing images forged by the combination of the four editing patterns and two editing methods based on approximately 3,500 real receipt images that we collected.

\begin{table}[hbtp]
    \caption{Statistics of FD-VIED. The number of images for each case in the training and test partitions.}
    \label{tabale:statistics}
    \centering
    \begin{tabular}{|c|c|c|c|}
        \hline
        &            & \textbf{training} & \textbf{test} \\ 
        \hhline{|====|}
        \multirow{2}{*}{Text removal} & Copy--move & 5796 & 728 \\
        \hhline{|~---|}
        & DNN--based & 6122 & 798 \\ 
        \hhline{|----|}
        \multirow{3}{*}{Text replacement} & Copy--move & 3019 & 394 \\ 
        \hhline{|~---|}
        & Mix\footnotemark[1]        & 6088 & 790 \\ 
        \hhline{|~---|}
        & DNN--based & 3061 & 399 \\ 
        \hhline{|----|}
        \multirow{2}{*}{Background addition} & Copy--move & 5598 & 704 \\ 
        \hhline{|~---|}
        & DNN--based & 5598 & 704 \\ 
        \hhline{|----|}
        \multirow{2}{*}{Text addition} & Copy--move & 5796 & 728 \\ 
        \hhline{|~---|}
        & DNN--based & 5796 & 728 \\ 
        \hhline{|====|}
        & total & 46874 & 5973 \\ 
        \hline
    \end{tabular}
    \footnotetext[1]{
    Mix refers to another implementation of the text-replacement pattern that combines blank pasting via inpainting and string pasting via the copy--move method.
    }
\end{table}

\section{Experiments}
This section is structured as follows.
Implementation details are described in Section \ref{ImplementationDetails}.
The exploration of the optimal pre-training method for document forgery detection is presented in Section \ref{exam:pre-train}. The evaluation and comparison of the proposed method and traditional forgery detection method are presented in Section \ref{section:finetune}.
The ablation study is presented in Section \ref{abulation}.
Finally, we examine the practicality of our model in Section \ref{wild}.

\begin{table*}[tbp]
  \caption{Evaluation results of document forgery detection. Each model is pre-trained on a different dataset and fine-tuned on FD-VIED.}
  \label{table:data_type}
  \centering
  \begin{tabular}{|l|l|r|r|}
    \hline
    Dataset for pre-training & Image Type & Image Num  &  mIoU  \\
    \hline 
    (No pre-trained)     & -                    & 0      & 0.667 \\
    RVL-CDIP           & Document             & 400k   & 0.828 \\
    ImageNet Subset    & Natural              & 400k   & 0.835 \\
    ImageNet           & Natural              & 1.28M  & 0.850 \\
    ImageNet+RVL-CDIP  & Natural and Document & 1.68M  &  0.857 \\
    \hline
  \end{tabular}
\end{table*}

\subsection{Implementation Details}
\label{ImplementationDetails}

\subsubsection{Pre-training stage}
We use SimCLR as a pre-training framework and two different datasets, namely ImageNet \cite{ImageNet}, which consists of natural images, and RVL-CDIP \cite{RVL-CDIP}, which consists of document images.
The resolution of input images is 224$\times$224.
ResNet50 \cite{7780459} is used as the base encoder framework for feature extraction.
The projection head outputs a 128-dimension vector from a 2048-dimension encoder output through a 2048-dimension hidden layer.
We use a batch size $N=256$ and LARS (with a learning rate of 0.3, weight decay of 1e-06, and momentum of 0.9) \cite{https://doi.org/10.48550/arxiv.1708.03888} as the optimizer.
Data augmentation for SimCLR is performed using RandomResizedCrop, RandomHorizonalFlip, Colorjitter (20\%), RandomGrayscale (20\%), and GaussianBlur (50\%).
We use the temperature of the loss $\tau = 0.1$. We pre-train the model for 200 epochs on two graphical processing units (NVIDIA A100-SXM4-40GB). The training takes 18 hours per 100,000 images.

\subsubsection{Fine-tuning stage} 
We use FCN \cite{7298965} for the downstream task and fine-tune it through supervised learning without freezing.
The input of the model is RGB images and the output of the model is pixel-wise predictions of the two-class (authentic/forged) categorization.
We use the cross-entropy loss, a batch size $N=8$, and the stochastic gradient descent algorithm (with a learning rate of 0.01, momentum of 0.9, and weight decay of 0.0005) \cite{https://doi.org/10.48550/arxiv.1609.04747} as the optimizer.
Input images are augmented by adopting Resize (1024$\times$1024), RandomCrop (512$\times$512), RandomFlip (50\%), and Photometric distortion. The model trains for 40,000 iterations on the same graphical processing units in the pre-training stage. The training takes approximately 24 hours. 

\subsubsection{Evaluation} 
After pre-training and fine-tuning, the models are evaluated for document forgery detection using FD-VIED. Almost all images in FD-VIED have very few forged areas compared with authentic areas. To account for this imbalance, we used the mean intersection over union (mIoU) as an evaluation metric. Each experiment is conducted three times and the average mIoU score is calculated.

\begin{table*}[hbtp]
  \caption{
  Evaluation results on document forgery detection by the mean IoU of the two classes (mIoU). Please note that, given the very small proportion of forged areas, if no forged areas are detected at all, the mIoU will be approximately 0.5.
  }
  \label{table:baseline}
        \centering
        \begin{tabular}{|c|c|c|c|c|c|c|}
        \hline
        \multicolumn{2}{|c|}{} & Ours & Traditional & NEDB-Net & MVSS-Net\\
            \hline
            \multirow{2}{*}{Text removal} 
            & Copy--move & 0.831 & 0.851 & 0.384 & 0.498  \\
            & DNN--based & 0.862 & 0.624 & 0.380 & 0.499 \\
            \hline
            \multirow{3}{*}{Text replacement} 
            & Copy--move & 0.783 & 0.823 & 0.384 & 0.501 \\
            & Mix       & 0.835  & 0.766 & 0.390 & 0.497 \\
            & DNN--based & 0.845 & 0.728 & 0.382 & 0.497 \\  
            \hline
            \multirow{2}{*}{
            \begin{tabular}{c} Background\\ addition
            \end{tabular}}
            & Copy--move & 0.809 & 0.832 & 0.379 & 0.499 \\
            & DNN--based & 0.853 & 0.606 & 0.378 & 0.501\\ 
            \hline    
            \multirow{2}{*}{Text addition} 
            & Copy--move & 0.845 & 0.836 & 0.382 & 0.498 \\
            & DNN--based & 0.843 & 0.803 & 0.382 & 0.497 \\ 
            \hline 
            \multicolumn{2}{|c|}{Average} & \textbf{0.834} & 0.763 & 0.382 & 0.499 \\
            \hline
        \end{tabular}
\end{table*}

\subsection{Exploring the Optimal Pre-training Method}
\label{exam:pre-train}
To explore the optimal pre-training method for document forgery detection, we pre-train each model on different datasets, RVL-CDIP (400k), ImageNet Subset (400k), ImageNet (1.28M), and ImageNet (1.28M) + RVL-CDIP (400k). Note that ImageNet (1.28M) is the original ImageNet and ImageNet Subset (400k) is a subset of it.

Table \ref{table:data_type} gives the evaluation results after fine-tuning for document forgery detection. All models with pre-training outperform the model without it, this means that pre-training is also effective for the document forgery detection task. The model pre-trained on ImageNet (1.28M) + RVL-CDIP (400k) has the best performance.

The model performing best for data augmentation in SimCLR involves RandomResizedCrop, RandomHorizonalFlip, Colorjitter, RandomGrayscale, and GaussianBlur augmentations. Excluding some of these augmentations or adding Rotation or Text-Ovelay did not improve the model performance. It is considered that there are ineffective augmentations for document forgery detection.

\paragraph{Discussion}
The models pre-trained with RVL-CDIP and with ImageNet Subset have similar scores (RVL-CDIP: 0.828 and ImageNet Subset: 0.835 in Table \ref{table:data_type}).
This indicates that document images and natural images are equally useful in pre-training for document forgery detection, even though they are very different in composition.
Considering that a larger dataset makes pre-training more effective, it would be an effective pre-training strategy to prepare a larger dataset by merging datasets like ImageNet and RVL-CDIP regardless of consistency differences between natural and document images.
In support of this, the model performance is improved by 0.007 when adding RVL-CDIP to ImageNet (ImageNet: 0.850 and ImageNet+RVL-CDIP: 0.857 in Table \ref{table:data_type}).

\subsection{Model Evaluation and Comparison}
\label{section:finetune}
We compare the proposed method and a traditional forgery detection method. 
As a proposed method, we fine-tuned a model with a subset of FD-VIED containing forged images by all editing methods. 
As a traditional method, we fine-tuned the same model with a subset of FD-VIED that exclusively included forged images by copy--move method, namely excluding forged images by DNN--based methods. 
In both cases, we used a pre-trained model on ImageNet (1.28M) + RVL-CDIP (400k) and each subset contains 10,000 images. 
In addition, as part of our investigation, we also compared publicly trained models of NEDB-Net and MVSS-Net, which are state-of-the-art methods for forgery detection in natural images. Table \ref{table:baseline} shows the results.

\paragraph{Discussion 1.}
The traditional model, trained solely on forged images by copy--move methods, performed relatively poorly when applied to forged images by DNN--based methods, particularly in cases involving text removal and background addition (i.e., inpainting). On the other hand, the proposed model, trained on forged images by DNN--based methods as well, demonstrated good performance on such forged images. It is considered that detecting forged document images created by copy--move methods and DNN--based methods are distinct in terms of the clues. Consequently, it is essential to train models on a dataset of forged document images created by DNN-based methods as proposed in our study.

\paragraph{Discussion 2.}
Our model performed well for each forgery pattern. Considering that text addition only adds text without changing the background, and that the text removal and the background pasting occur in empty background areas without text, this result suggests that our model uses inconsistent backgrounds and fonts as clues for document forgery detection. 

\paragraph{Discussion 3.}
NEDB-Net and MVSS-Net scored less than the chance rate in all cases. Even when the hyperparameters were adjusted, they tended to either output no region as forged areas or classify the entire document region as forged, the latter is the reason why the mIoU of NEDB-Net dropped even lower than 0.5 in Table \ref{table:baseline}.
It is considered that detecting forged document images and detecting forged natural images are essentially different in terms of methods and clues.

\begin{table*}
  \captionsetup{singlelinecheck=off, justification=raggedright}
  \caption{
  Evaluation results for document forgery detection (mIoU) when each model is fine-tuned with a different dataset.
  }
  \label{table:abulation1}
  \setlength{\tabcolsep}{3pt}
  \centering
  \scalebox{0.84}{%
    \begin{tabular}{|c|c|c|c|c|c|c|c|c|c|c|c|c|c|c|c|}
    \hhline{|*{16}{-|}}
    \multicolumn{2}{|c|}{} & \multicolumn{2}{c|}{Case 1} & \multicolumn{2}{c|}{Case 2} & \multicolumn{2}{c|}{Case 3} & \multicolumn{2}{c|}{Case 4} & \multicolumn{2}{c|}{Case 5} & \multicolumn{2}{c|}{Case 6} & \multicolumn{2}{c|}{Case 7} \\
    \hhline{|*{16}{-|}}
    \multicolumn{2}{|c|}{} & Train & Test & Train & Test & Train & Test & Train & Test & Train & Test & Train & Test & Train & Test \\ 
    \hline
    \multirow{2}{*}{\begin{tabular}[c]{@{}c@{}}Text\\removal\end{tabular}}
    & Copy--move & $\checkmark$ & 0.831 &  & 0.774 & $\checkmark$ & 0.835 & $\checkmark$ & 0.835 & $\checkmark$ & 0.842 & $\checkmark$ & 0.851 & & 0.512 \\
    & DNN--based & $\checkmark$ & 0.862 & & 0.828 & $\checkmark$ & 0.866 & $\checkmark$ & 0.849 & $\checkmark$ & 0.867 & & 0.624 & $\checkmark$ & 0.872 \\ 
    \hline
    \multirow{3}{*}{\begin{tabular}[c]{@{}c@{}}Text\\replacement\end{tabular}}
    & Copy--move & $\checkmark$ & 0.783 & $\checkmark$ & 0.792 &  & 0.576  & $\checkmark$ & 0.805 & $\checkmark$ & 0.797& $\checkmark$ & 0.823 & & 0.517 \\
    & Mix & $\checkmark$ & 0.835 & $\checkmark$ & 0.839 &  & 0.778  & $\checkmark$ & 0.842 & $\checkmark$ & 0.835&  & 0.766 &  & 0.775 \\
    & DNN--based & $\checkmark$ & 0.845 & $\checkmark$ & 0.846 &  & 0.791  & $\checkmark$ & 0.852 & $\checkmark$ & 0.854 &  & 0.728 & $\checkmark$ & 0.862 \\  
    \hline
    \multirow{2}{*}{\begin{tabular}[c]{@{}c@{}}Background\\addition\end{tabular}}
    & Copy--move & $\checkmark$ & 0.809 & $\checkmark$ & 0.802 & $\checkmark$ & 0.809  &  & 0.750 & $\checkmark$ & 0.813 & $\checkmark$ & 0.832 & & 0.505 \\
    & DNN--based & $\checkmark$ & 0.853 & $\checkmark$ & 0.847 & $\checkmark$ & 0.857  &  & 0.801 & $\checkmark$ & 0.868 &  & 0.606 & $\checkmark$ & 0.868 \\ 
    \hline
    \multirow{2}{*}{\begin{tabular}[c]{@{}c@{}}Text\\addition\end{tabular}}
    & Copy--move & $\checkmark$ & 0.845 & $\checkmark$ & 0.853 & $\checkmark$ & 0.841  & $\checkmark$ & 0.844 &  & 0.712& $\checkmark$& 0.836 &  & 0.633 \\
    & DNN--based & $\checkmark$ & 0.843 & $\checkmark$ & 0.851 & $\checkmark$ & 0.838  & $\checkmark$ & 0.849 &  & 0.781&  & 0.803 & $\checkmark$ & 0.849 \\ 
    \hline
    \multicolumn{2}{|c|}{Average} &  & \textbf{0.834} & & 0.826 & & 0.799 & & 0.825 & & 0.817 & & 0.763 & & 0.736 \\
    \hhline{|*{16}{-|}}
    \end{tabular}
    \captionsetup{justification=centering}
    }
\end{table*}

\subsection{Ablation Study}
\label{abulation}
To analyze the effectiveness of training each forgery case, we fine-tune models with different subsets of FD-VIED. Each subset contains 10,000 images but differs in the editing pattern or method as shown in Table \ref{table:abulation1}.

The result suggests the value of training models with a wider range of forged image variations. In particular, as shown in cases 6 and 7, not only does the model's performance suffer on images forged by DNN-based methods when trained exclusively on images forged by copy-move methods, but the reverse scenario is also predominantly true.
So, it is crucial to incorporate diverse variations, especially editing methods such as copy--move and DNN--based rather than editing patterns.

\subsection{Performance testing in the wild}
\label{wild}
\begin{figure*}
    \centering
    \begin{tabular}{ccc}
        \toprule
         Original image & Forged image & Detection result
         \\ 
        \midrule
        \includegraphics[width=3.5cm]{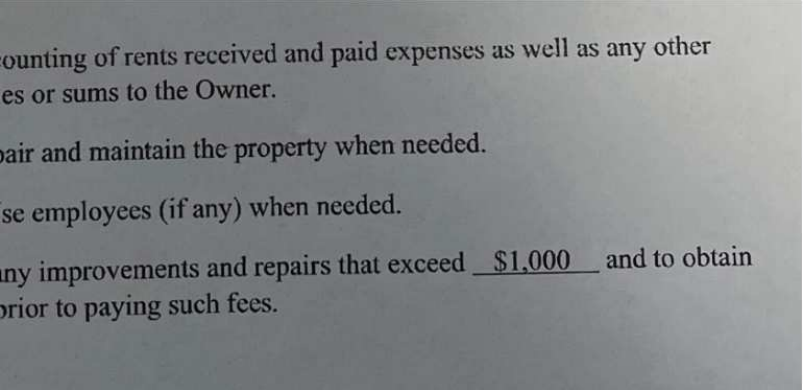} &
        \includegraphics[width=3.5cm]{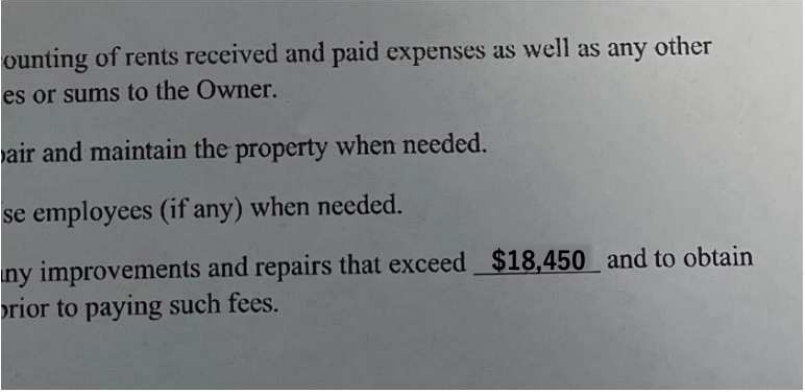} &
        \includegraphics[width=3.5cm]{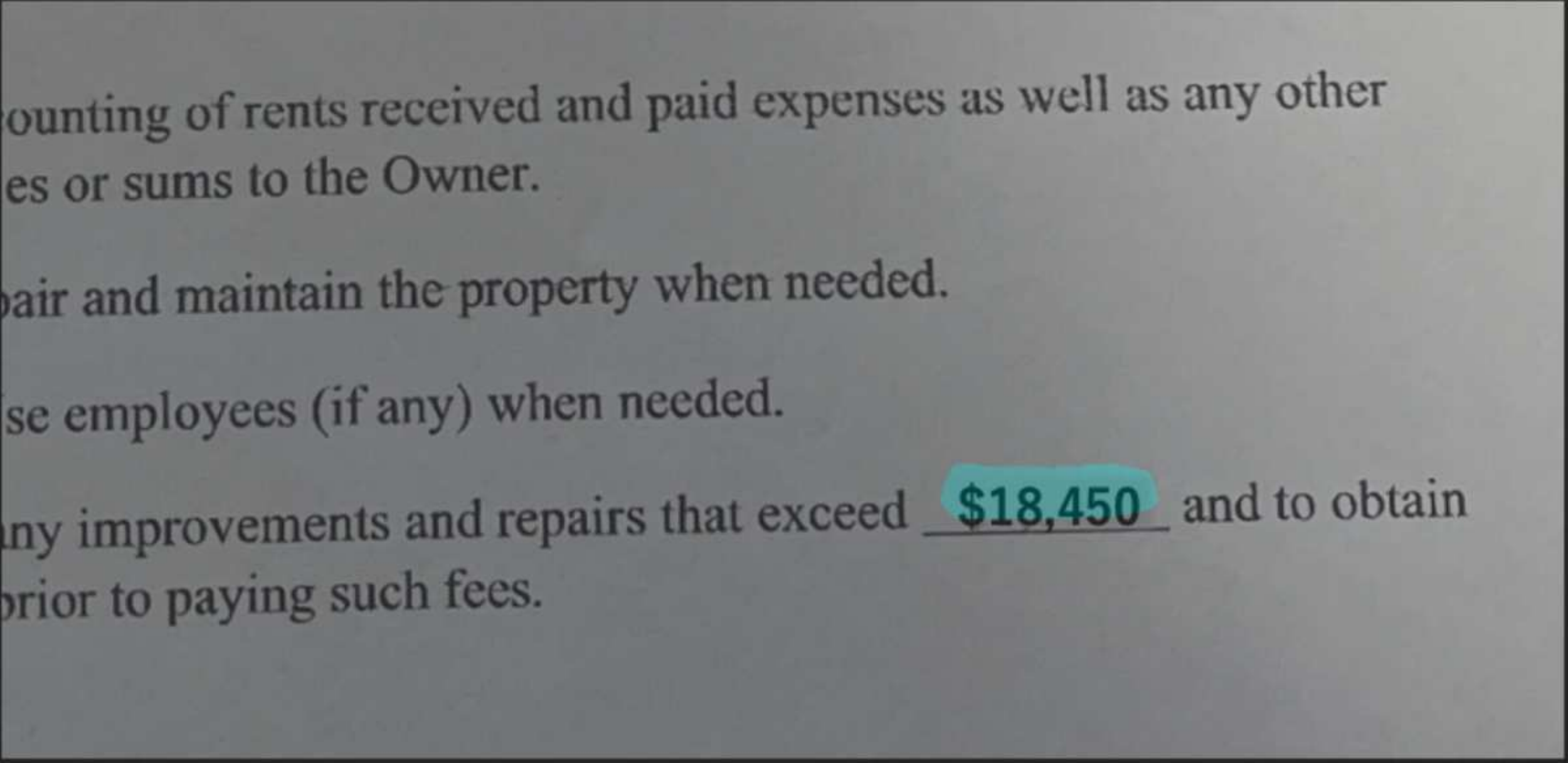}
        \\ 
        \includegraphics[width=3.5cm]{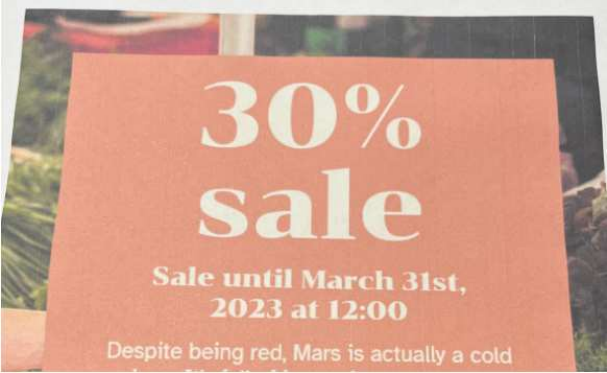} &
        \includegraphics[width=3.5cm]{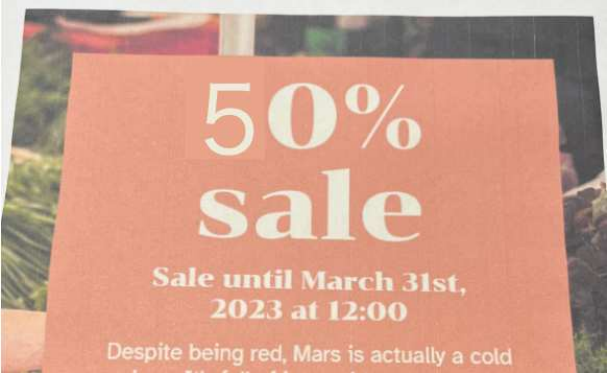} &
        \includegraphics[width=3.5cm]{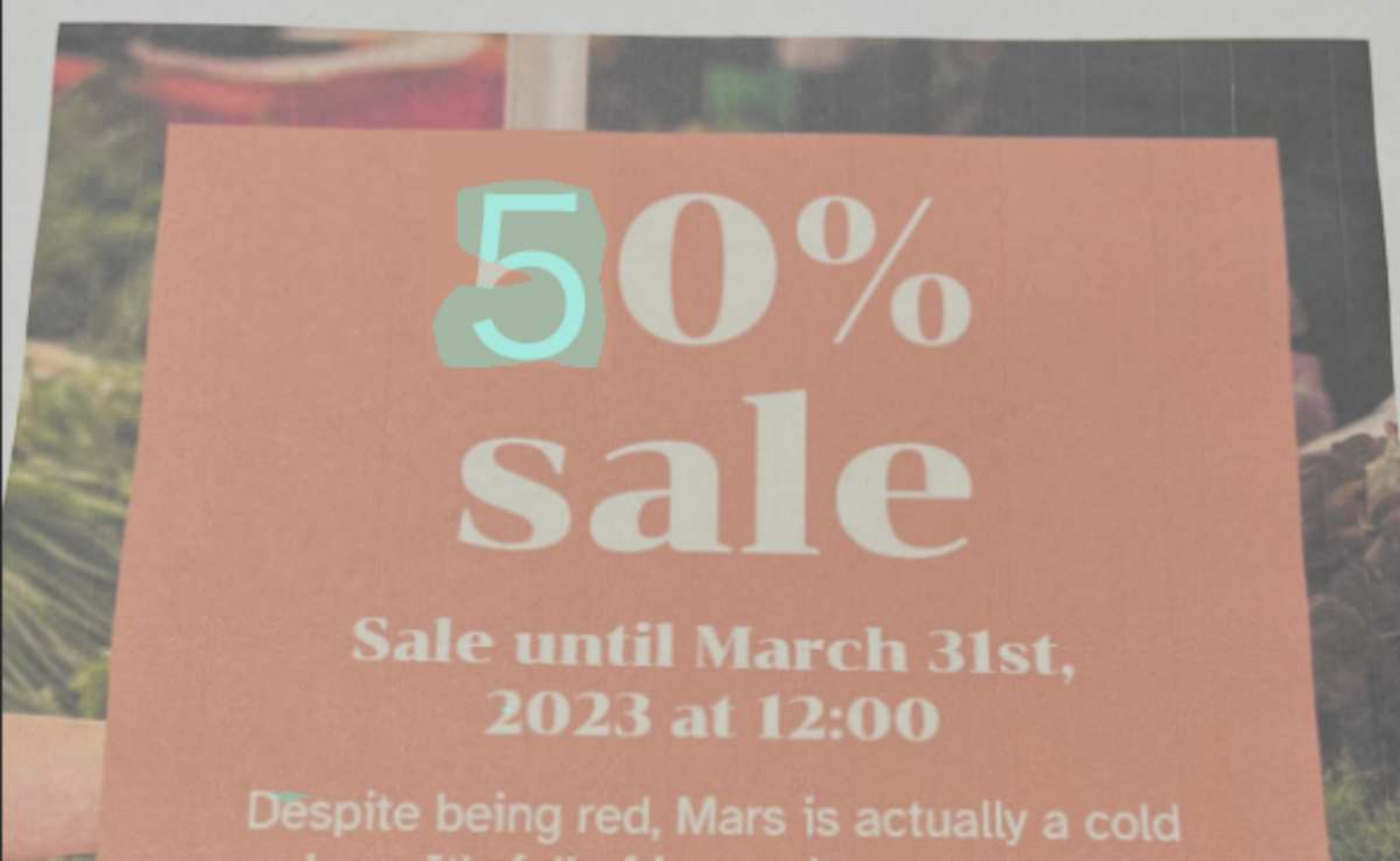}
        \\ 
        \includegraphics[width=3.5cm]{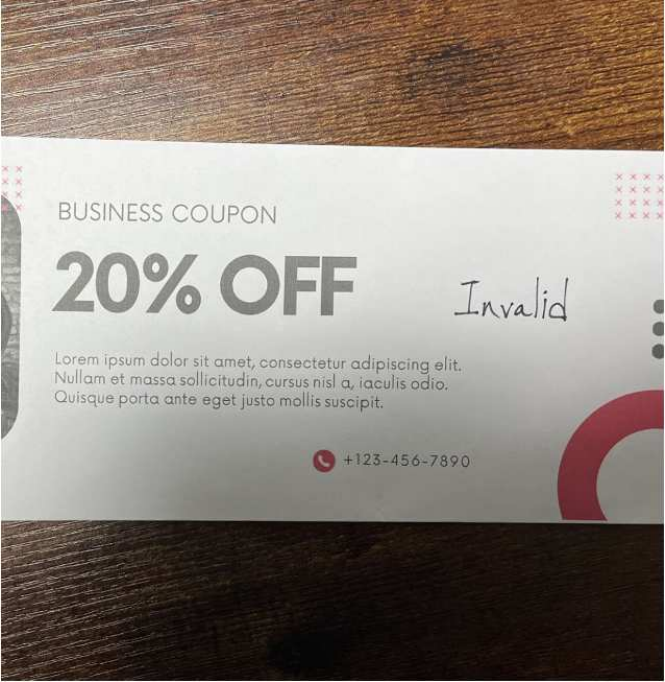} &
        \includegraphics[width=3.5cm]{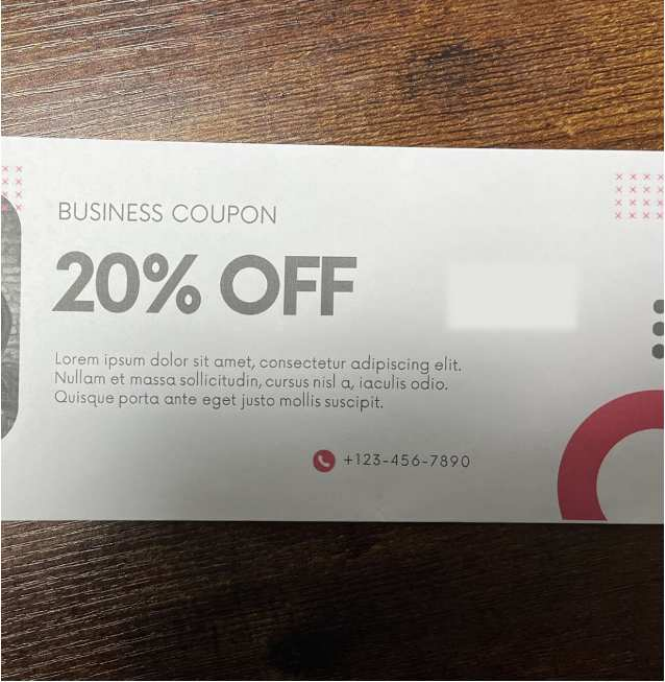} &
        \includegraphics[width=3.5cm]{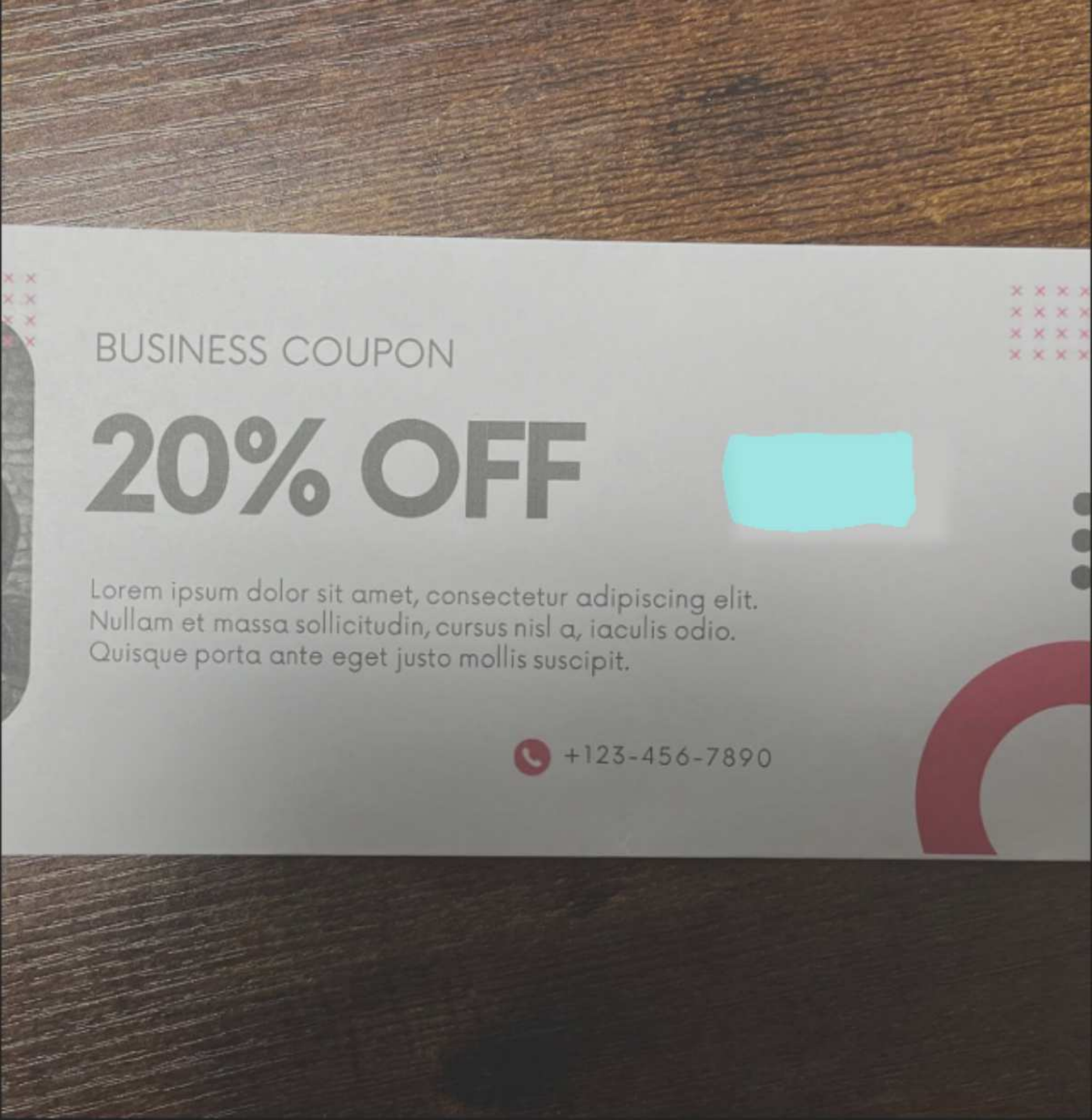}
        \\ 
        \includegraphics[width=3.5cm]{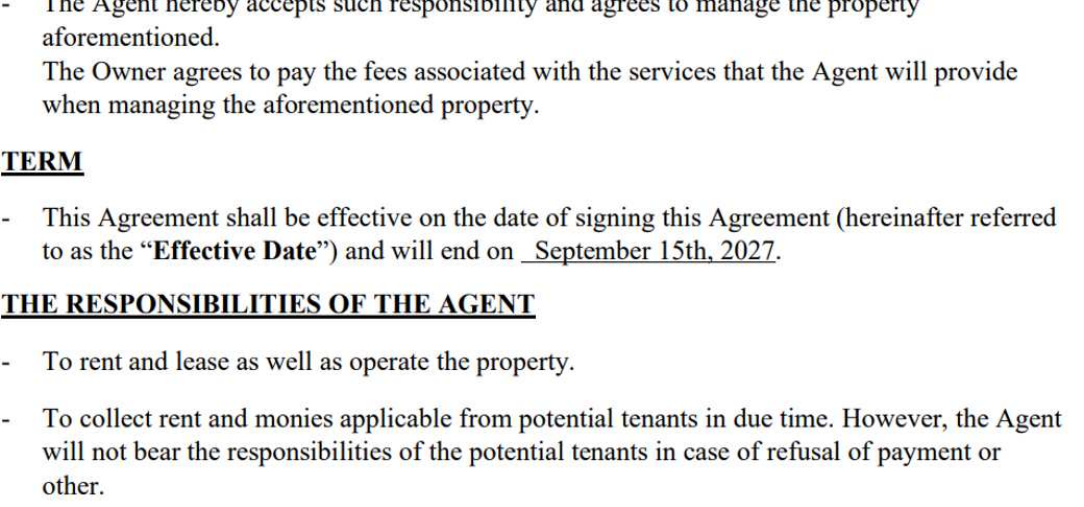} &
        \includegraphics[width=3.5cm]{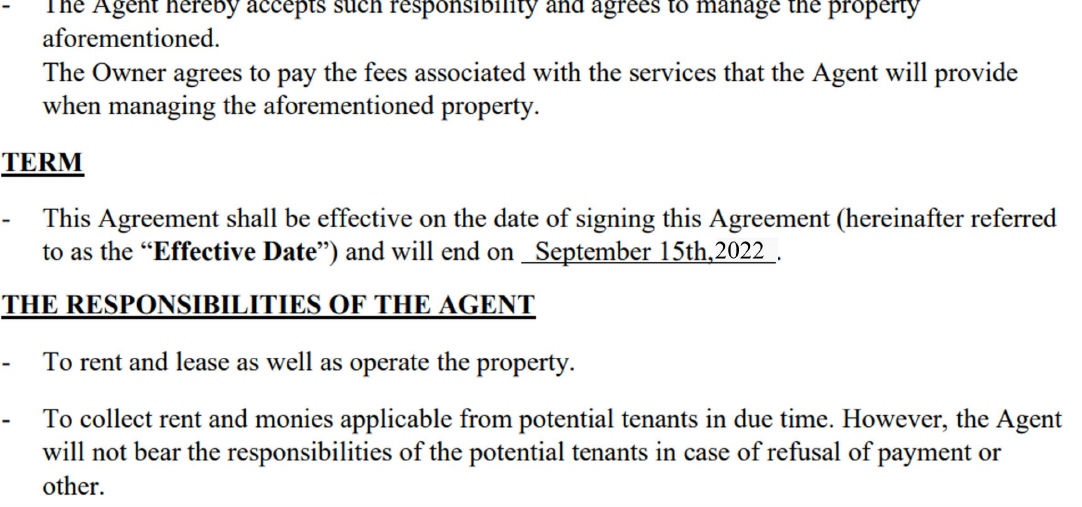} &
        \includegraphics[width=3.5cm]{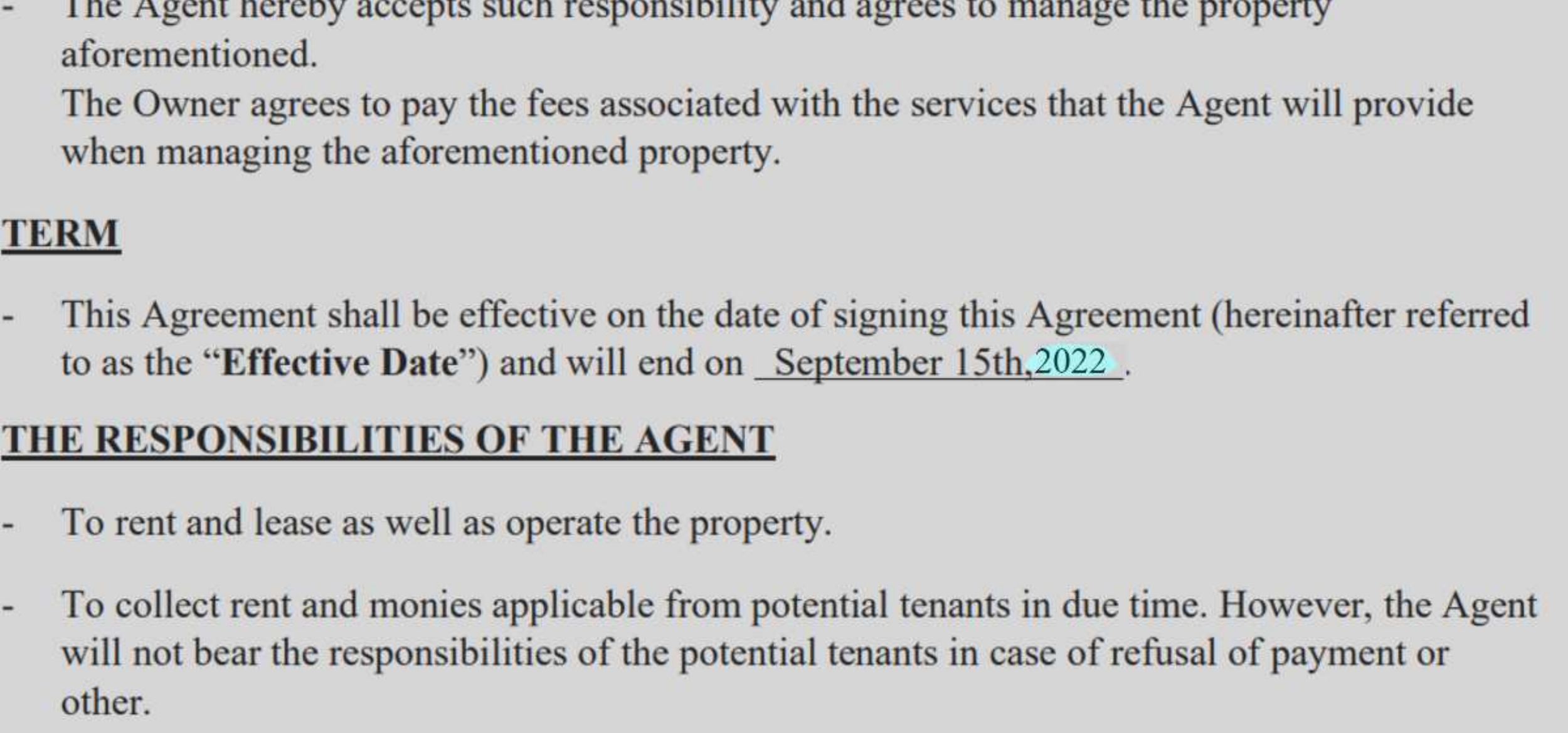}
        \\ 
        \includegraphics[width=3.5cm]{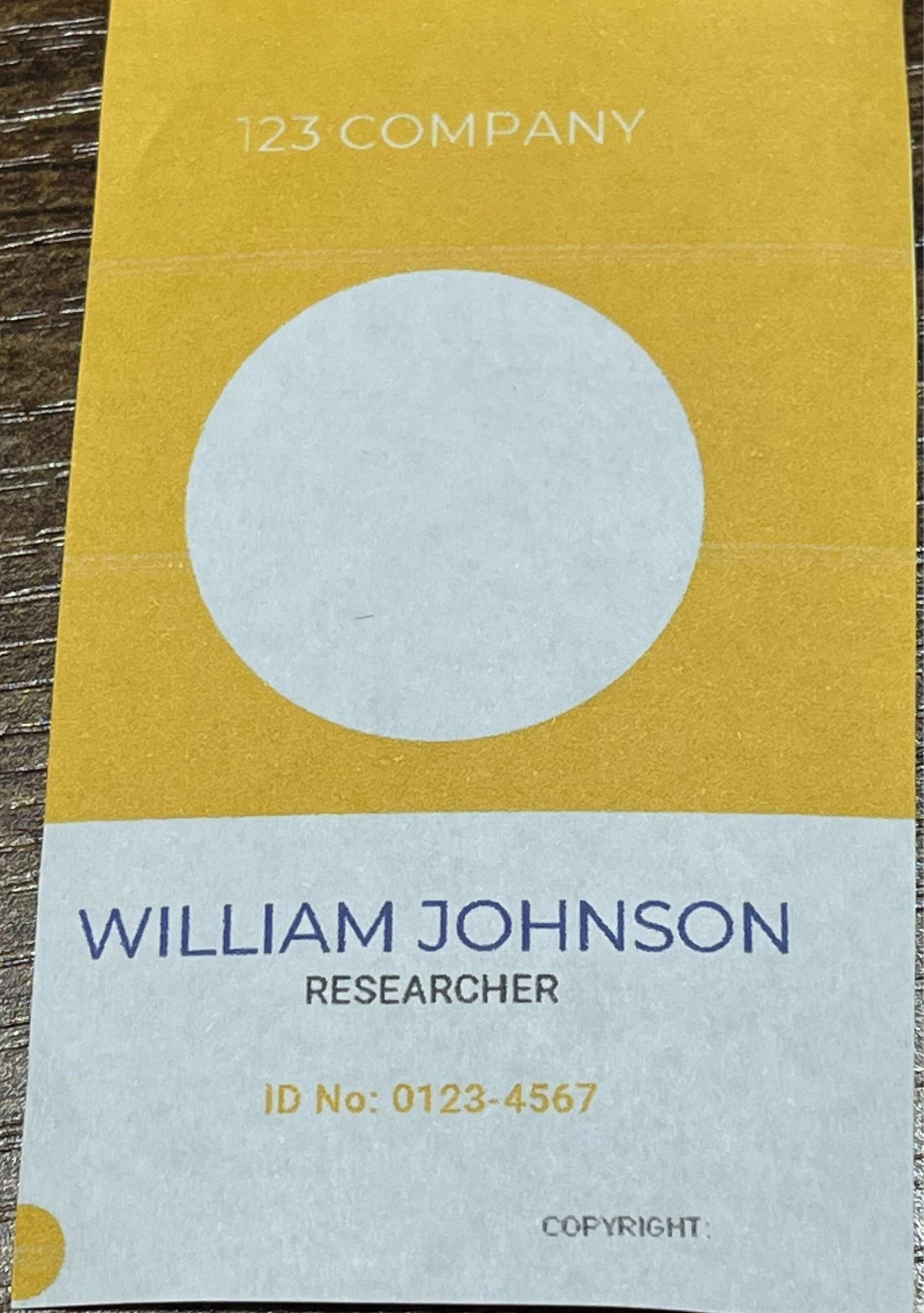} &
        \includegraphics[width=3.5cm]{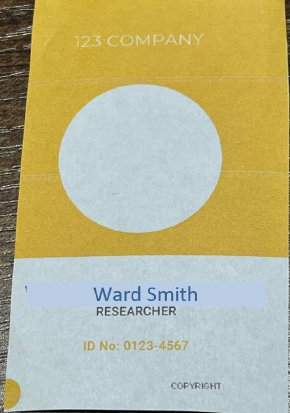} &
        \includegraphics[width=3.5cm]{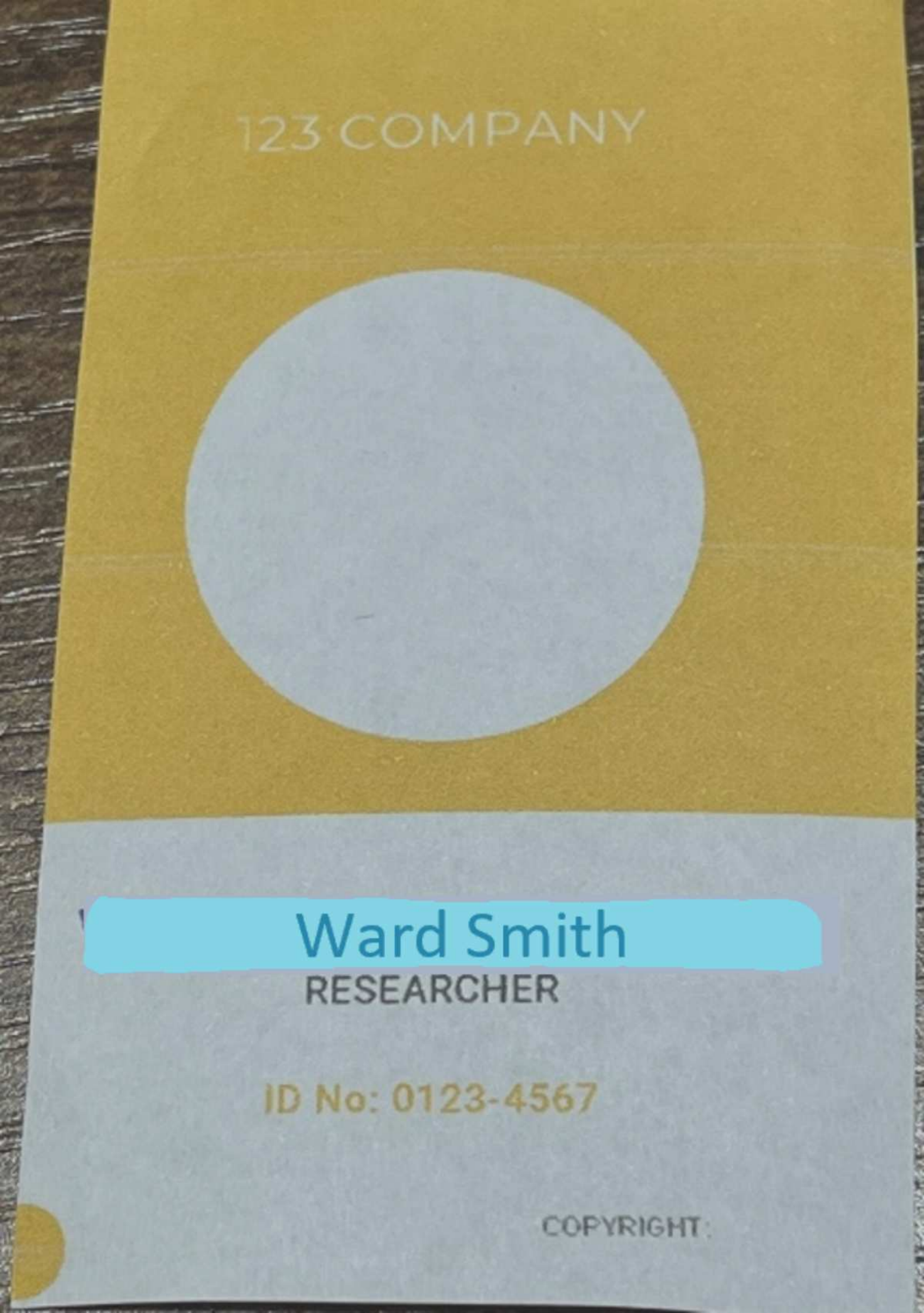}
        \\ 
        \includegraphics[width=3.5cm]{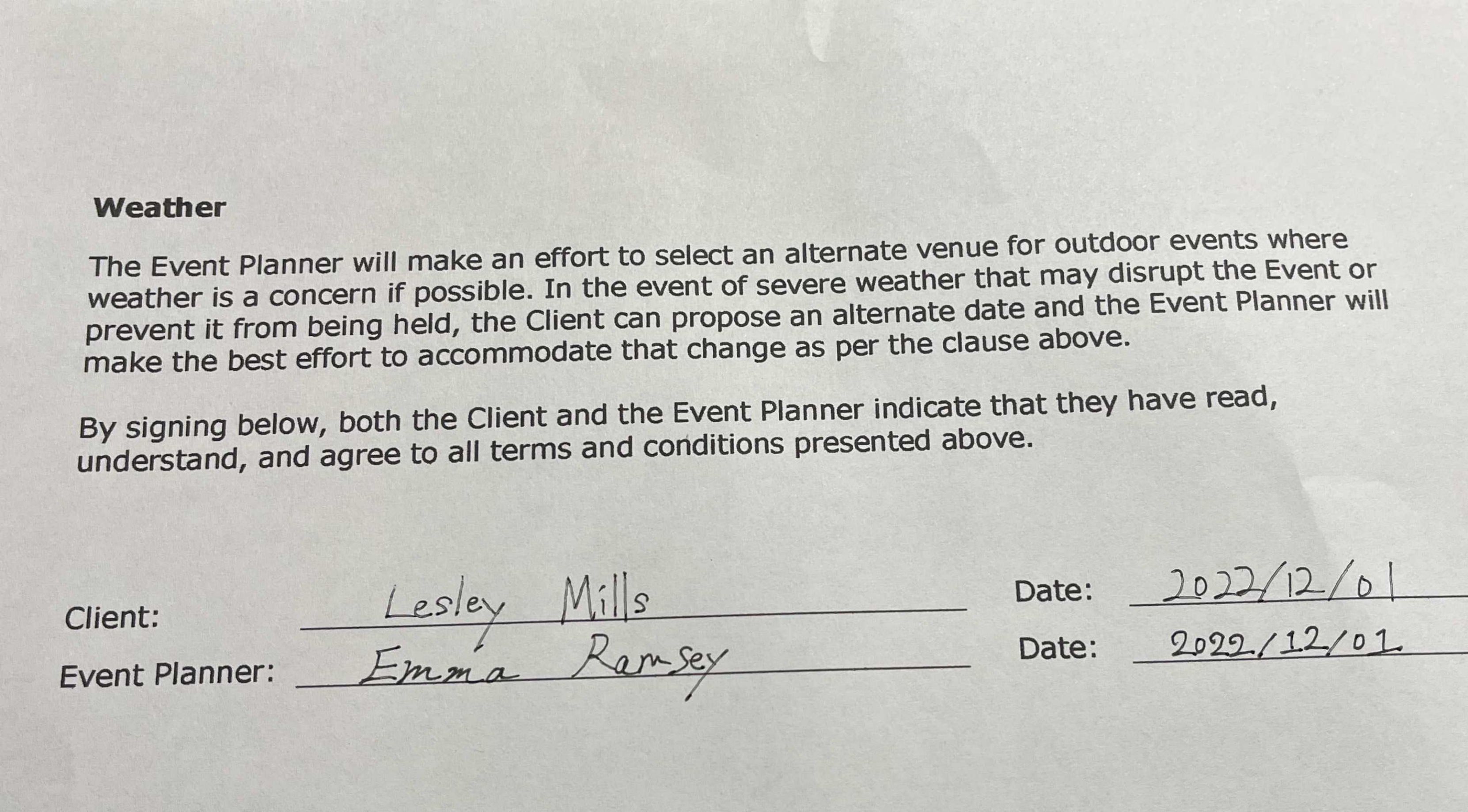} &
        \includegraphics[width=3.5cm]{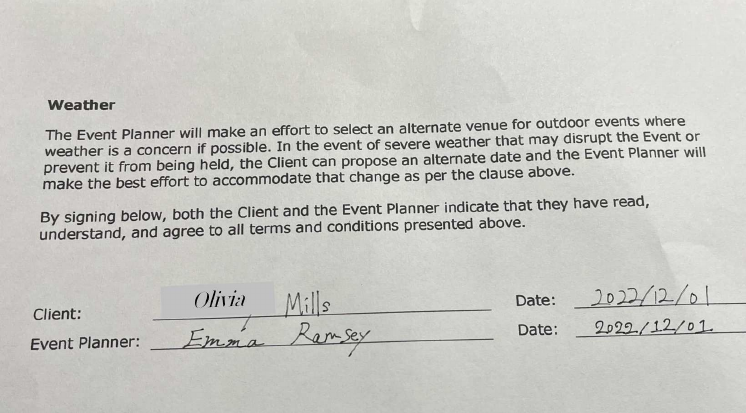} &
        \includegraphics[width=3.5cm]{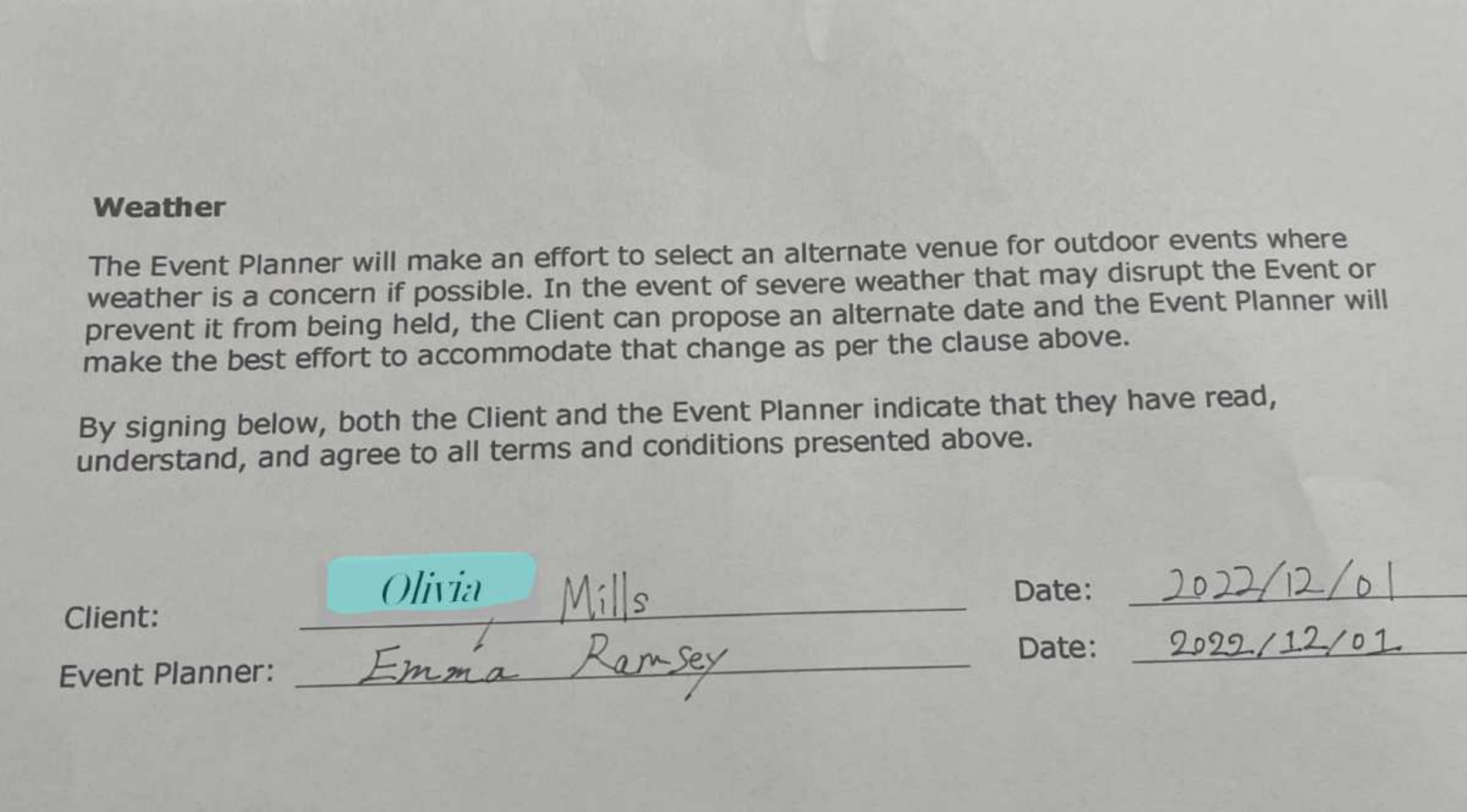}
        \\ 
        \\
    \bottomrule 
    \end{tabular}
    \caption{Example output of our document forgery detection model. Our model performs well even for contracts and coupons that are not included in FD-VIED. (The areas highlighted in blue are inferred as edited.)}
    \label{fig:DMDsample}
\end{figure*}

To confirm whether the proposed model is sufficiently useful and performs well in practical use, we recruited participants to create forged document images using various software and methods of their choice.
These images included not only receipts but also contracts and coupons, which are not included as document types in FD-VIED. Figure\ \ref{fig:DMDsample} shows some results. The model performed well regardless of differences in the document type and editing process.

\section{Conclusion}
The recent advancements in DNN methods for generative tasks may amplify the threat of forged documents. To address this, our work 1) proposed a dataset named FD-VIED containing document images forged with a wide variety including DNN--based methods, 2) devised an effective pre-training strategy for document forgery detection, and 3) demonstrated the effectiveness of our approach in detecting forged documents created by DNN--based methods. We hope that this study will contribute to future developments in this field.

\bibliographystyle{abbrv}
\bibliography{sn-bibliography}


\end{document}